\documentclass{article}

\usepackage{natbib}
\usepackage{microtype}
\usepackage{graphicx}
\usepackage{subfigure}
\usepackage{booktabs} 

\usepackage[pdfborder={0 0 1}]{hyperref}
\let\svthefootnote\thefootnote
\newcommand\freefootnote[1]{%
  \let\thefootnote\relax%
  \footnotetext{#1}%
  \let\thefootnote\svthefootnote%
}

\usepackage{amsmath}
\usepackage{amssymb}
\usepackage{mathtools}
\usepackage{amsthm}

\usepackage[capitalize,noabbrev]{cleveref}

\theoremstyle{plain}
\newtheorem{theorem}{Theorem}[section]

\newtheorem{lemma}[theorem]{Lemma}

\theoremstyle{definition}

\newtheorem{assumption}[theorem]{Assumption}
\theoremstyle{remark}

\usepackage[textsize=tiny]{todonotes}

\usepackage{graphicx} 
\usepackage{color}
\usepackage{amsmath}
\usepackage{mathtools}
\usepackage{amssymb}
\usepackage{amsthm}
\usepackage{appendix}
\usepackage{cleveref}
\newtheorem{fact}{Fact}

\newtheorem*{lemma*}{Lemma}

\usepackage{enumitem}

\newcommand{\sj}[1]{{\color{green}[SJ:#1]}}
\newcommand{\khasha}[1]{{\color{blue}{khasha:#1}}}
\newcommand{\khashared}[1]{{#1}}
\newcommand{\trace}{\mathrm{Tr}}
\newcommand{\hess}{D^2 \mathcal L}
\newcommand{\phiprime}{\phi'(\theta_j^\top x_i)}
\newcommand{\phitwoprime}{\phi''(\theta_j^\top x_i)}
\newcommand{\phithreeprime}{\phi'''(\theta_j^\top x_i)}
\newcommand{\NN}{\textsf{NN}}

\newcommand\numberthis{\addtocounter{equation}{1}\tag{\theequation}}
\newcommand{\thetazero}{\theta_0}
\newcommand{\barthetazero}{{\bar\theta}_0}
\newcommand{\thetastarj}{\theta^*_j}
\newcommand{\thetastar}{\theta^*}

\newcommand{\loss}{\mathcal L}
\newcommand{\Rmd}{\mathbb R^{md}}
\newcommand{\stack}[1]{\Big[#1\Big]}

\newcommand{\tracehess}{\trace\hess}
\newcommand{\Rr}{\mathbb R^{md}}
\newcommand{\manifold}{\mathcal M}
\newcommand{\hypersurface}{\mathcal M}
\newcommand{\nproj}{P^N_{\theta}}
\newcommand{\thetaproj}{P_\theta}
\newcommand{\brackets}[1]{\left\{#1\right\}}
\newcommand{\tangentspace}[1]{\mathcal T_{#1}(\manifold)}
\newcommand{\normalspace}[1]{\mathcal T^N_{#1}(\manifold)}
\newcommand{\phii}{\psi}

\usepackage{arxiv}

\usepackage[utf8]{inputenc} 
\usepackage[T1]{fontenc}    
\usepackage{url}            
\usepackage{booktabs}       
\usepackage{amsfonts}       
\usepackage{nicefrac}       
\usepackage{microtype}      
\usepackage{lipsum}
\usepackage{graphicx}
\graphicspath{ {./images/} }

\title{Simplicity Bias via Global Convergence of Sharpness Minimization}

\author{
 Khashayar Gatmiry \\
  MIT \\
  \texttt{gatmiry@mit.edu} \\
   \And
 Zhiyuan Li \\
  TTIC \\
  \texttt{zhiyuanli@ttic.edu} \\
  \And
 Sashank J. Reddi \\
  Google Research \\
  \texttt{sashank@google.edu} \\
  \And
  Stefanie Jegelka\\
  $\text{TUM}^{*} \text{and} 
 ~\text{MIT}^{\dagger}$\\
  \texttt{stefje@mit.edu}\\
}

\renewcommand{\thefootnote}{\fnsymbol{footnote}}
\begin{document}
\maketitle
\footnotetext[1]{CIT, MCML, MDSI}
\footnotetext[2]{EECS and CSAIL}
\begin{abstract}
The remarkable generalization ability of neural networks is usually attributed to the implicit bias of SGD, which often yields models with lower complexity using simpler (e.g. linear) and low-rank features~\cite{huh2021low}. Recent works have provided empirical and theoretical evidence for 
the bias of particular variants of SGD (such as label noise SGD) toward flatter regions of the loss landscape. 
Despite the folklore intuition that flat solutions are 'simple', the connection with the simplicity of the final trained model (e.g. low-rank) is not well understood. 
In this work, we take a step toward bridging this gap by studying the \emph{simplicity structure} that arises from minimizers of the sharpness for a class of two-layer neural networks.
We show that, for any high dimensional training data and certain activations, with small enough step size, label noise SGD always converges to a network that replicates a single linear feature across all neurons; thereby, implying a simple 
rank one feature matrix. To obtain this result, our main technical contribution is to show that label noise SGD always minimizes the sharpness on the manifold of models with zero loss for two-layer networks. 
Along the way, we discover a novel property --- a local geodesic convexity --- of the trace of Hessian of the loss at approximate stationary points on the manifold of zero loss, which links sharpness to the geometry of the manifold. This tool may be of independent interest. 
\end{abstract}


\section{Introduction}
    Overparameterized neural networks trained by stochastic gradient descent (SGD) have demonstrated remarkable generalization ability. The emergence of this ability, even when the network perfectly fits the data and without any explicit regularization, still remains a mystery~\cite{zhang2017understanding}. 
A line of recent works have attempted to provide a theoretical basis for such observations by showing that \emph{SGD has an implicit bias toward functions with ``low complexity" among all possible networks with zero training loss}~\citep{neyshabur2017exploring, soudry2018implicit,gunasekar2017implicit,li2018algorithmic,gunasekar2018characterizing,woodworth2020kernel,li2019towards, haochen2021shape,pesme2021implicit}. In particular, it is noted that SGD tends to pick simpler features to fit the data over more complex ones   ~\citep{hermann2020shapes,kalimeris2019sgd,neyshabur2014search,pezeshki2021gradient}. This tendency is typically referred to as \emph{simplicity bias}.
     Notably, \citet{huh2021low} empirically observes that the feature matrix on the training set tends to become low-rank and that this property is amplified in deeper networks. 
    
    On the other hand, recently, a different form of implicit bias of SGD based on the sharpness of loss landscape has garnered interest. The study of flatness of the loss landscape and its positive correlation with generalization is not new (e.g.~\citep{hochreiter1997flat}), and has been supported by a plethora of empirical and theoretical evidence~\citep{keskar2016large,dziugaite2017computing,jastrzkebski2017three,neyshabur2017exploring,wu2018sgd,jiang2019fantastic,blanc2020implicit,wei2019data,wei2022statistically,haochen2021shape,foret2020sharpness,damian2021label,li2021happens,ma2021linear,ding2022flat,nacson2022implicit,wei2022statistically,lyu2022understanding,norton2021diametrical}. It has been suggested that, after reaching zero loss, neural networks trained using some variants of SGD are biased toward regions of the loss landscape with low sharpness~\citep{blanc2020implicit,damian2021label,li2021happens,arora2022understanding,lyu2022understanding,li2022fast,wen2022does,bartlett2022dynamics}. 
    In particular, \citet{blanc2020implicit} discovered the sharpness-reduction implicit bias of a variant of SGD called \emph{label noise SGD}, which explicitly adds noise to the labels. They proved that SGD locally diverges from points that are not stationary for the trace of the Hessian of the loss after fitting the training data. Along this line, \citet{li2021happens} show that in the limit of step size going to zero, label noise SGD converges to a gradient flow according to the trace of Hessian of the loss after reaching almost zero loss. This flow can be viewed as a Riemannian gradient flow with respect to the sharpness on the manifold of zero loss, \emph{i.e.}, the flow following the  gradient of sharpness projected onto the tangent space of the manifold. 
    Furthermore, a standard Bayesian argument based on minimum description length suggests such flat minimizers correspond to ``simple'' models.
    
    From a conceptual point of view, both simplicity bias and sharpness minimization bias suggest that SGD learns \emph{simple} solutions; however, the nature of "simplicity" is different. Simplicity bias is defined in the feature space, meaning the model learns simple features, \emph{e.g.}, low-rank features, but low sharpness is defined in the parameter space, meaning the loss landscape around the learned parameter is flat, such that the model is resilient to random parameter perturbations and admits a short and simple description from a Bayesian perspective. It is thus an interesting question whether these two notions of ``simplicity''  for neural networks have any causal relationship. 
    Existing works that attempt to study this connection are restricted to somewhat contrived settings (e.g. \cite{blanc2020implicit,shallue2018measuring,szegedy2016rethinking,shallue2018measuring,haochen2021shape}). For instance, \citet{blanc2020implicit} study 1-D two-layer ReLU networks and two-layer sigmoid networks trained on a single data point, where they show the stationary points of the trace of Hessian on the manifold of zero loss are solutions that are simple in nature. They further prove that label noise SGD locally diverges from non-stationary points.  As another example, \citet{li2021happens} \khashared{and~\citet{vivien2022label} independently} show that label noise SGD can recover the sparse ground truth, but only for a simple overparameterized quadratic linear model.
    In this paper, we study the following fundamental question:
    \begin{quote}
    \emph{(1) Is there a non-linear general setting 
    where the sharpness-reduction implicit bias \emph{provably} implies simplicity bias?}
    \end{quote}

\begin{table*}[t]
\centering
\begin{tabular}{l| l| l}
Results  & Assumptions & Basic Description\\
\hline
\hline
Theorem~\ref{thm:stationary}
& \begin{tabular}{@{}l@{}} $\phi', \phi''' > 0$ \text{or} $\phi = x^3$\\
$X$ is full-rank \end{tabular} & \begin{tabular}{@{}l@{}} The global minimizer of the squared loss, \\
all neurons collapse into the same vector in the data subspace\end{tabular}\\
\hline
Theorem~\ref{thm:rateofconvergenceone}
& \begin{tabular}{@{}l@{}} $\phi', \phi''' > 0$\\
$X^\top X \geq \mu I$\\ $\beta$-normality~[\ref{assump:three}] \end{tabular} & \begin{tabular}{@{}l@{}} The Riemannian gradient flow of sharpness converges  \\
 to the global minimizer of sharpness on the manifold of zero loss\end{tabular}\\
\hline
Theorem~\ref{thm:main_intro}
& \begin{tabular}{@{}l@{}} $\phi', \phi''' > 0$ \text{or} $\phi = x^3$\\
$X^\top X \geq \mu I$\\ $\beta$-normality~[\ref{assump:three}] \end{tabular} & \begin{tabular}{@{}l@{}} label noise SGD with small enough step size \\
converges to the global minimizer of the sharpness on the manifold of zero loss\end{tabular}
\end{tabular}
\caption{Summary of main theoretical results and key assumptions}
\label{table:summary}
\vspace{-1.2em}
\end{table*}
    
However, even if we can show that a model with minimum flatness is a simple model, a key open question is whether 
the sharpness-reduction implicit bias, characterized by the Riemannian gradient flow of sharpness in \citep{li2021happens}, even successfully minimizes the sharpness in the first place. The convergence of the gradient flow so far is only known in very simple settings, \emph{i.e.}, quadratically overparametrized linear nets~\cite{li2021happens}.
      In this regard, a central open question in the study of sharpness minimization for label noise SGD is:
    \begin{quote}
    \emph{(2) Does label noise SGD converge to the global minimizers of the sharpness on the manifold of zero loss, and if it does, how fast does it converge? In particular, for what class of activations does global convergence happen?
    }  
    \end{quote}
    In this paper, we show sharpness regularization provably leads to simplicity bias for two-layer neural networks with certain activation functions. We further prove that the  Riemannian gradient flow on the manifold linearly converges to the global minimizer of the sharpness \emph{for all initialization}. To our best knowledge, this is the first global linear convergence result for the gradient flow of trace of Hessian on manifolds of minimizers.

More formally, we consider the mean squared loss 
\begin{align}
    \loss(\theta) = \sum\nolimits_{i=1}^n (\sum\nolimits_{j=1}^m \phi(\theta_j^\top x_i) - y_i)^2\label{eq:meansquaredloss}   
\end{align}
 for a two-layer network model where the weights of the second layer are fixed to one and $\{x_i,y_i\}_{i=1}^n$ are the training dataset.~\footnote{We discuss how our result generalizes to the case of unequal weights at the end of  Section~\ref{sec:stationarysketch}.} \khashared{We consider label noise SGD on $\loss$, that is, running gradient descent on the loss $\sum_{i=1}^n (\sum_{j=1}^m \phi(\theta_j^\top x_i)-y_i + \zeta_{t,i})^2$ at each step $t$, in which independent Gaussian noise $\zeta_{t,i} \sim \mathcal N(0, \sigma^2)$ is added to the labels.}  The following is a corollary of the main results of our paper, which proves the asymptotic convergence of label noise SGD to the global minimizer of the sharpness. This result holds under Assumptions~\ref{assump:one},~\ref{assump:two},~\ref{assump:three}) and for sufficiently small learning rate. 
\begin{theorem}[Convergence to simple solutions from any initialization]\label{thm:main_intro}
    Given any $\epsilon>0$ and arbitrary initial parameter $\theta[0]$, for label noise SGD with \khashared{any noise variance $\sigma^2>0$} initialized at $\theta[0]$, there exists a threshold $\eta_0$ such that for any learning rate $\eta \leq \eta_0$, running label noise SGD with step size $\eta$ and number of iterations \khashared{$T= \tilde O(\log(1/\epsilon)\cdot \eta^{-2}\sigma^{-2})$}, with probability at least $1-\epsilon$,
\begin{enumerate}
    \item $\loss(\theta[T])\le \epsilon$;
    \item $\mathrm{Tr}(D^2\loss(\theta[T])) 
    \le \inf_{\theta:\loss(\theta)=0}\mathrm{Tr}(D^2\loss(\theta)) + \epsilon$;
    \item For all neurons ${\theta}_j[T]$ and all data points $x_i$, 
        $\big|{{\theta}_j[T]}^\top x_i - \phi^{-1}(y_i/m)\big| \leq \epsilon,$ 
\end{enumerate}
where $D^2 \loss$ is the Hessian of the loss and we hide constants about loss $\loss$ and initialization $\theta[0]$ in $O(\cdot)$.
\end{theorem}

\begin{figure}[ht]
\vskip -0.1in   
\begin{center}
\centerline{\includegraphics[width=0.5\columnwidth]{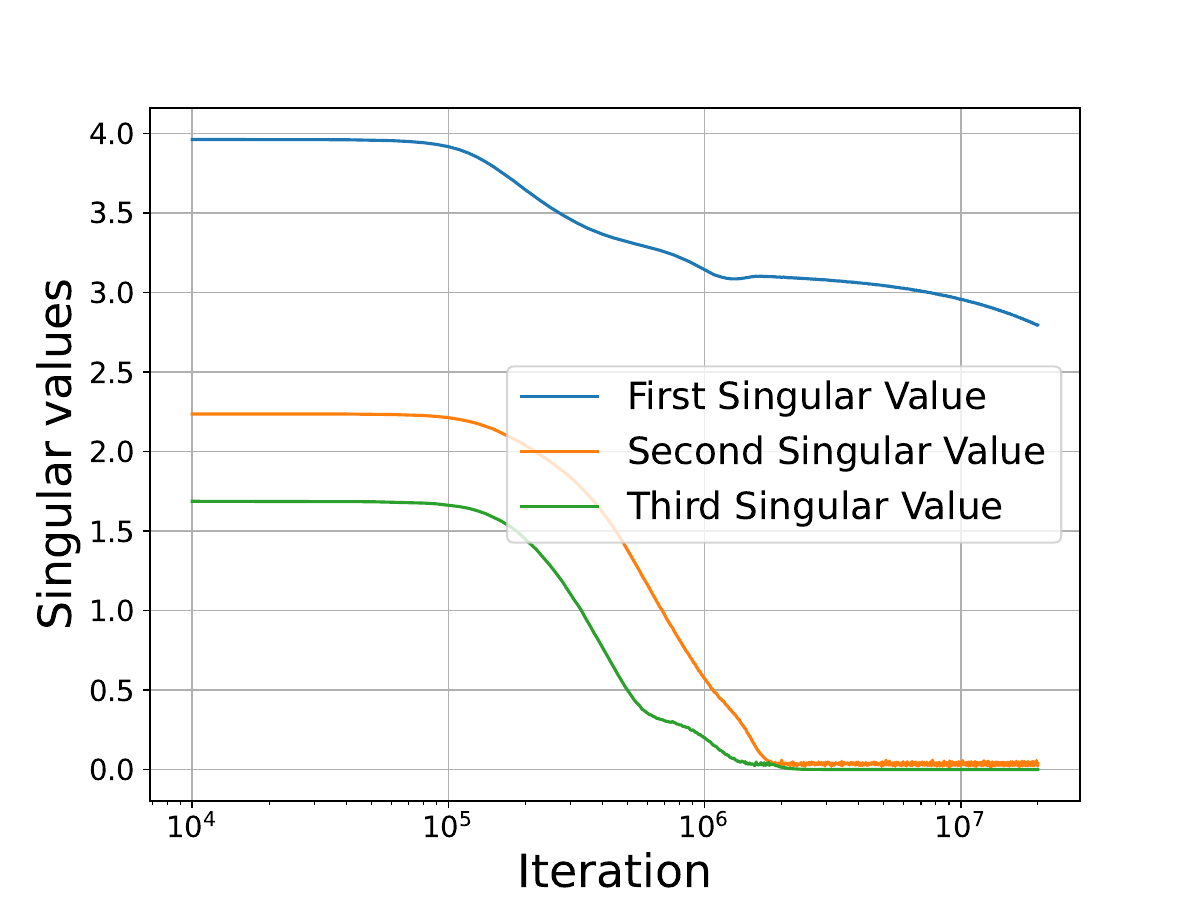}}
\caption{The rank of the feature and weight matrices go to one as Theorem~\ref{thm:main_intro} predicts in the high dimensional setting $d > n$.\\ \\}
\label{fig:one}
\end{center}
\vskip -0.4in
\end{figure}

 Informally, ~\Cref{thm:main_intro} states that, first, label noise SGD reaches zero loss and then (1) successfully minimizes the sharpness to its global optima (2) and more importantly, it recovers the simplest possible solution that fits the data, in which the 
     feature matrix is rank one (see Appendix~\ref{sec:proofthm2} for proof details). In particular, Theorem~\ref{thm:main_intro} holds for pure label noise SGD without any additional projection.  
\begin{itemize}[leftmargin=*]
    \item {\bf Simplicity bias:}
    We prove that at every minimizer of the trace of Hessian, the pre-activation of different neurons becomes the same for every training data.
    In other words, all of the feature embeddings of the dataset with respect to different neurons collapse into a single vector, \emph{e.g.}, see \Cref{fig:one}. 
    This combined with our convergence result mathematically proves that the low-rank feature observation by~\citep{huh2021low} holds in our two-layer network setting trained by sharpness minimization induced regularizer algorithms, such as label noise SGD and 1-SAM.

    \item {\bf Convergence analysis:}
 Under an additional normality assumption on the activation~(\Cref{assump:two}), we show that the gradient flow of the trace of Hessian converges exponentially fast to the global optima on the manifold of zero loss, and as a result, label noise SGD with a sufficiently small step size successfully converges to the global minimizer of trace of Hessian among all the models achieving zero loss. Importantly, we do not assume additional technical conditions (such as PL inequality and non-strict saddle property used in prior works) about the landscape of the loss or its sharpness, and our constants are explicit and only depend on the choice of the activation and coherence of the data matrix. 
 Moreover, our convergence results hold in the strong sense i.e., the convergence holds for the last iterations of label noise SGD.

We have summarized our main results in Table~\ref{table:summary}. The novelty of our approach (after that we show it converges to zero loss) is that we characterize the convergence on the manifold of zero loss in two phases (1) the first phase where the algorithm is far from stationary points, trace of Hessian decreases \khashared{with a at least constant rate} and there is no convexity \khashared{in trace of Hessian}, (2) the second phase where the algorithm gets close to stationarity, we prove a novel g-convexity property of trace of Hessian which holds only locally on the manifold, but we show implies an exponentially fast convergence rate by changing the Lyapunov potential from value of trace of Hessian to the norm squared of its gradient on the manifold. We further prove that approximate stationary points are close to a global optimum via a semi-monotonicity property.
\end{itemize}

\begin{figure}[ht]
\vskip 0.2in
\begin{center}
\centerline{\includegraphics[width=0.5\columnwidth]{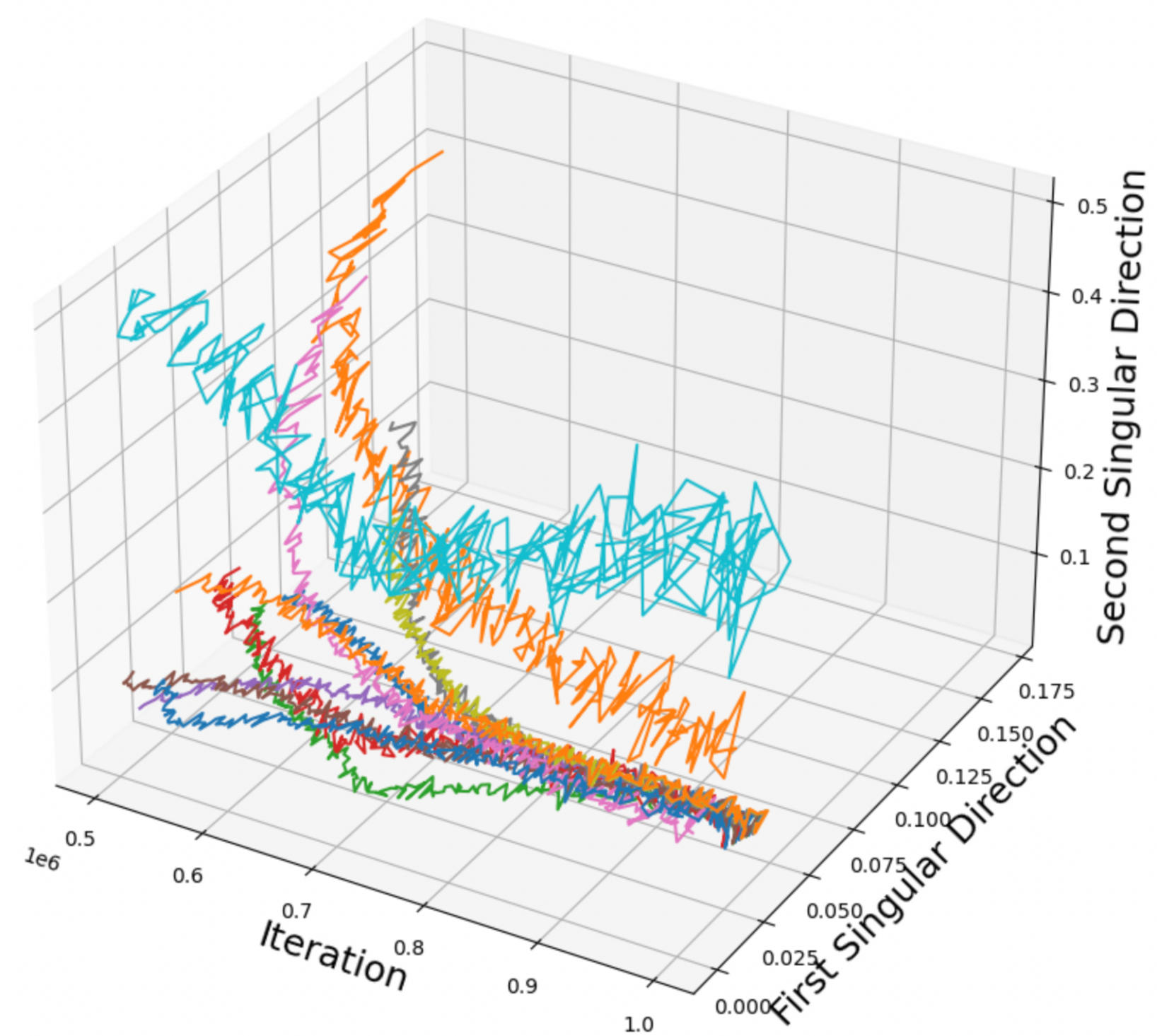}}
\caption{The feature embeddings of neurons collapse into clusters in the low dimensional regime $d < n$. The plot visualizes this phenomenon using the first and second principal components of the feature embedding $\Theta^* X$.}
\label{fig:two}
\end{center}
\vskip -0.2in
\end{figure}

Interestingly, we observe this simplicity bias even beyond the regime of our theory; namely when the number of data points exceeds the ambient dimension, instead of a single vector, the feature embeddings of the neurons cluster and collapse into a small set of vectors (see \Cref{fig:two}).

\section{Related Work}

    \paragraph{Implicit Bias of Sharpness Minimization.}  Recent theoretical investigations, including those by \citet{blanc2019implicit}, \citet{damian2021label}, and \citet{li2021happens}, have indicated that Stochastic Gradient Descent (SGD) with label noise intrinsically favors local minimizers with smaller Hessian traces. This is under the assumption that these minimizers are connected locally as a manifold \khashared{and these analyses focus on the final phase of training when iterates are close to the manifold of minimizes. The effect of label noise or mini-batch noise in the central phase of training, \emph{i.e.}, when the loss is still substantially above zero is more difficult and is only known for simple quadratically overparametrized linear models~\citep{vivien2022label,andriushchenko2023sgd}.} Further, \citet{arora2022understanding} demonstrated that normalized Gradient Descent (GD) intrinsically penalizes the Hessian's largest eigenvalue. \citet{ma2021linear} proposes that such sharpness reduction phenomena could also be triggered by a multi-scale loss landscape. In the context of scale-invariant loss functions, \citet{lyu2022understanding} found that GD with weight decay implicitly reduces the spherical sharpness, defined as the Hessian's largest eigenvalue evaluated at the normalized parameter.

Another line of research examines the sharpness minimization effect of large learning rates, assuming that (stochastic) gradient descent converges at the end of training. This analysis primarily hinges on the concept of linear stability, as referenced in works by \citet{wu2018sgd}, \citet{cohen2021gradient}, \citet{ma2021linear}, and \citet{cohen2022adaptive}. More recent theoretical analyses, such as those by \citet{damian2022self} and \citet{li2022analyzing}, suggest that the sharpness minimization effect of large learning rates in gradient descent does not necessarily depend on the convergence assumption and linear stability. Instead, they propose a four-phase characterization of the dynamics in the so-called Edge of Stability regime, as detailed in \citet{cohen2021gradient}.

 Sharpness-aware Minimization (SAM, \citet{foret2020sharpness}) is a highly effective regularization method that improves generalization by penalizing the sharpness of the landscape. SAM was independently developed in \citep{zheng2021regularizing,norton2021diametrical}.  Recently, \citet{wen2022does,bartlett2022dynamics} proved that SAM successfully minimizes the worst-direction sharpness under the assumption that the minimizers of training loss connect as a smooth manifold.~\citet{wen2022does} also analyze SAM with batch size 1 for regression problems and show that the implicit bias of SAM becomes penalizing average-direction sharpness, which is approximately equal to the trace of the Hessian of the training 
loss. \khashared{\citet{andriushchenko2023sharpness} observe that the SAM update rule for ReLU networks is locally biased toward sparsifying the features leading to low-rank features. However, their analysis is local and does not provide a convergence proof. They further confirm empirically that deep networks trained by SAM are biased to produce low-rank features.}

\section{Main Results}

{\bf Problem Setup.}
In this paper, we focus on the following two-layer neural network model:
\begin{align}
    r_{\theta, \NN}(x) \triangleq \sum_{j=1}^m \phi(\theta_j^\top x).\label{eq:basicmodel}
\end{align}
Here $\theta = (\theta_1,\dots,\theta_m)$ and $\phi$ are the set of parameters and activation of the neural network, respectively. As illustrated in Equation~\eqref{eq:basicmodel}, we assume that all the second-layer weights are equal to one. \khashared{At the end of section~\ref{sec:stationarysketch} we discuss how our approach in Theorem~\ref{thm:stationary} generalizes to any fixed choice of weights in the second layer.}
We  denote the matrix of the weights by $\mathrm \Theta \in {\mathbb R}^{m\times d}$, whose $j^{\text{th}}$ row is $\theta_j$. Given a training dataset $\brackets{(x_i, y_i)}_{i=1}^n$,  
 we study the mean squared loss
\begin{align}
    \loss(\theta) \triangleq \sum_{i=1}^n (r_{\theta, \NN}(x_i) - y_i)^2.\label{eq:meansquaredloss}
\end{align}
For simplicity, we assume that the data points have unit norm, i.e. $\forall i\in [n]$, $\|x_i\| = 1$. Let $\manifold$ denote the zero level set of $\loss$: $\manifold \triangleq \brackets{\theta \in \Rr \big| \ \loss(\theta) = 0}$.
As we show in Lemma~\ref{lem:manifoldwelldefined} in the Appendix, $\mathcal M$ is well-defined as a manifold due to a mild non-degeneracy condition implied by Assumption~\ref{assump:one}. We equip $\mathcal M$ with the standard Euclidean metric in $\mathbb{R}^{md}$.

The starting point of our investigation is the work \cite{li2021happens}, which characterizes the minimizer of the trace of the Hessian of the loss, $\tracehess$, 
 as the implicit bias of label noise SGD by showing that, in the limit of step size going to zero, label noise SGD evolves on the manifold of zero loss according to the following deterministic gradient flow:
\begin{align}
    \frac{d}{dt}\theta(t) \triangleq -\nabla \trace\hess(\theta(t)).\label{eq:gradientflow}
\end{align}
Here, 
$\nabla$ is the Riemannian gradient operator on the manifold $\manifold$, which is the projection of the normal Euclidean gradient onto the tangent space of $\manifold$ at $\theta(t)$. 
\khashared{We use this result of~\cite{li2021happens} to relate the convergence of label noise SGD to the convergence of the flow~\eqref{eq:gradientflow}. Our main object of study in this paper is then the global convergence of this flow.}
Note that starting from a point $\theta(0)$ on $\manifold$, $\theta(t)$ remains on $\manifold$ for all times $t$ (for a quick recap on basic notions in Differential Geometry such as covariant derivative and gradient on a manifold, we refer the reader to Appendix~\ref{sec:rgeometry}.)
On a manifold, similar to Euclidean space, having a zero gradient $\|\nabla \tracehess(\thetastar)\| = 0$ (the gradient defined on the manifold) at some points $\theta^*$ is a necessary condition for being a local minimizer, or equivalently the projection of Euclidean gradient onto the tangent space at $\theta^*$ has to vanish. More generally, we call $\theta$ an $\epsilon$-stationary point on $\manifold$ if $\|\nabla \tracehess(\theta)\| \leq \epsilon$.

\subsection{Characterization of Stationary Points}

Before characterizing the stationary points of $\tracehess$, we introduce an important assumption on the activation function, 
namely the convexity and positivity of its derivative.
\begin{assumption}[Strict positivity and convexity of $\phi'$]\label{assump:one}
    For all $z \in \mathbb R$, $\phi'(z) \geq \varrho_1 >0, \phi'''(z) \geq \varrho_2>0$.
\end{assumption}
For example, Assumption~\ref{assump:one} is satisfied for $\phi(x) = x^{3} + \nu x$ with constants $\varrho_1 = \nu, \varrho_2 = 1$.  
In Lemma~\ref{lem:boundedregion}, given the condition $\phi''' > 0$ we show that $\theta(t)$ remains in a bounded domain along the gradient flow, which means having the weaker assumption that $\phi', \phi''' > 0$ automatically implies Assumption~\ref{assump:one} for some positive constants $\varrho_1$ and $\varrho_2$. 
Another activation of interest to us that happens to be extensively effective in NLP applications is the cube ($x^3$) activation~\cite{chen2014fast}. Although $x^3$ does not satisfy the strict positivity required in Assumption~\ref{assump:one}, we separately prove Theorem~\ref{thm:stationary} for $x^3$ (see Lemma~\ref{lem:cube} in the Appendix.)

 Next, let $X \triangleq \big(x_1\big|\dots\big| x_n\big)$ be the data matrix. We make a coherence assumption on $X$ which requires the input dimension $d$ to be at least as large as the number of data points, denoted by $n$. 
\begin{assumption}[Data matrix coherence]\label{assump:two}
     The data matrix $X$ satisfies
        $X^\top X \geq \mu I.$
\end{assumption}
Assumption~\ref{assump:one} implies that $\manifold$ is well-defined as a manifold 
(see Lemma~\ref{lem:manifoldwelldefined} for a proof).
Under aforementioned Assumptions~\ref{assump:one} and~\ref{assump:two}, we show that the trace of the Hessian regularizer has a unique stationary point on the manifold.
\begin{theorem}[First-order optimal points]\label{thm:stationary}
Under Assumptions~\ref{assump:one} and~\ref{assump:two}, the first order optimal points and global optimums of $\tracehess$ on $\manifold$ coincide and are equal to the set of all $\thetastar = \stack{\thetastarj}_{j=1}^m$ such that 
for all $i \in [n]$ and $j \in [m]$ satisfy the following:
\begin{align}
    {\thetastarj}^\top x_i = \phi^{-1}(y_i/m).\label{eq:effectofdepth}
\end{align}
\end{theorem}

Note that instead of Assumption~\ref{assump:two}, just having the linear independence of $\{x_i\}_{i=1}^n$ suffices to prove Theorem~\ref{thm:stationary}, as pointed out in section~\ref{sec:stationarysketch}.
In particular, Theorem~\ref{thm:stationary} means that the feature matrix is rank one at a global optimum $\theta^*$, which together with our main convergence result~\Cref{thm:main_intro} proves the low-rank bias conjecture proposed by~\citep{huh2021low} in the two-layer network setting for label noise SGD and 1-SAM.
Notice the $1/m$ effect of the number of neurons on the global minimizers of the implicit bias in Equation~\eqref{eq:effectofdepth}. 
While Theorem~\ref{thm:stationary} concerns networks with equal second-layer weights, the case of unequal weights can be analyzed similarly, as we discuss after the proof of Theorem~\ref{thm:stationary} in Section~\ref{sec:stationarysketch}.

\subsection{Convergence Rate}
Next, to state our convergence result, we introduce the key $\beta$-normality assumption, under which we bound the rate of convergence of the trace of Hessian to its global minimizer.
\begin{assumption}[$\beta$-normality]\label{assump:three}
    For all $z\in \mathbb R$ the second derivative of $\phi(z)$ can be bounded by the first and third derivatives as: 
    \begin{align*}
        \beta \phi''(z) \leq {\phi'}^2(z)\phi'''(z).\label{eq:normalactivation}
    \end{align*}
\end{assumption}

 An example class of activations that satisfies both Assumptions~\ref{assump:one} and~\ref{assump:three} is of the form $\phi(z) = z^{2k+1} + \nu z$ for $\nu > 0$, which is well-known in the deep learning theory literature~\cite{li2018algorithmic,jelassi2022vision,allen2020towards,woodworth2020kernel}. Notably, these activations satisfy Assumption~\ref{assump:three} with normality coefficient $\beta = \min\{\frac{1}{(2k+1)^2(2k-1)}, \nu^2\}$.

Under
Assumptions~\ref{assump:one},~\ref{assump:two}, and~\ref{assump:three}, we show that $\theta(t)$ converges to $\theta^*$ exponentially fast in Theorem~\ref{thm:rateofconvergenceone}. 
This results from strong local g-convexity of trace of Hessian for approximate stationary points (see Lemmas ~\ref{lem:psdness} and~\ref{lem:hessstrongconvexity}), plus a semi-monotonicity property for the trace of Hessian (Lemma~\ref{lem:closetoopt}).

\begin{theorem}[Convergence of the gradient flow]\label{thm:rateofconvergenceone}
Consider the limiting flow of Label noise SGD on the manifold of zero loss, which is the gradient flow in~\eqref{eq:gradientflow}.
    Then, under Assumptions~\ref{assump:one},\ref{assump:two}, and~\ref{assump:three},
    \begin{enumerate}[label=\textbf{(C.\arabic*)}]
        \item\label{claim:second} The gradient flow reaches $\epsilon$-stationarity for all times
    \[
    t\geq \frac{\tracehess(\theta(0))}{\mu \beta^2} + \frac{\log( \beta^2/(\varrho_1^2\varrho_2^2\epsilon^2) \vee 1)}{\varrho_1\varrho_2 \mu}.
    \]
    \item\label{claim:third} 
    For all $j \in [m]$ and $i \in [n]$, the dot product of the $j$th neuron to the $i$th data point becomes $\epsilon$-close to that of any global optimum $\theta^*$:
    \begin{align}
        \big|\theta_j(t)^\top x_i - {\theta^*_j}^\top x_i\big| \leq \epsilon,\label{eq:dotproductproximity}
    \end{align}
    where $\theta^*$ is defined in Theorem~\ref{thm:stationary}.
    \end{enumerate}
\end{theorem}
The formal proof of Theorem~\ref{thm:rateofconvergenceone} is given in Section~\ref{sec:proofthm2}. We sketch its proof in Section~\ref{sec:proofsketchthm3}. Note that the rate is logarithmic in the accuracy parameter $\epsilon$, i.e., the flow has exponentially fast convergence to the global minimizer. 

\section{Proof Sketches}\label{sec:proofsketches}
Before delving into the technical details of the proofs, we discuss some necessary background and  additional notation used in the proofs.
\subsection{Additional Notation}
 We use $f_i(\theta)$ to denote the output of the network on the $i$th input $x_i$, i.e.,
\begin{align}
    f_i(\theta) \triangleq r_{\theta, \NN}(x_i).\label{eq:twolayermodel}
\end{align}
 We use $f(\theta) \triangleq (f_1(\theta),\dots,f_n(\theta))$ to denote the array of outputs of the network on $\{x_i\}_{i=1}^n$. We use $D$ for the Euclidean directional derivative or Euclidean gradient, and $\nabla$ for the covariant derivative on the manifold or the gradient on the manifold. We denote the Jacobian of $f$ at point $\theta$ by $Df(\theta)$ whose $i$th row is $Df_i(\theta)$, the Euclidean gradient of $f_i$ at $\theta$. 
 Recall the definition of the manifold of zero loss as the zero level set of the loss $\loss$,
which is the intersection of zero level sets of functions $f_i, \forall i\in [m]$: $\manifold = \brackets{\theta\in \Rr| \ \forall i\in [n], \ f_i(\theta) = y_i}.$
 The tangent space $\tangentspace{\theta}$ of $\mathcal M$ at point $\theta$ can be identified by the tangents to all curves on $\manifold$ passing through $\theta$, and the normal space $\normalspace{\theta}$ in this setting is just the orthogonal subspace to $\tangentspace{\theta}$. We denote the projection operators onto $\tangentspace{\theta}$ and $\normalspace{\theta}$ by $\thetaproj$ and $\nproj$, respectively. 
For a set of vectors $\{v_i\}_{i=1}^n$ where for all $i\in [n]$, $v_i\in \mathbb R^d$, we use the notation $\stack{v_i}_{i=1}^n$ to denote the vector in $\mathbb R^{nd}$ obtained by stacking vectors $\{v_i\}_{i=1}^n$.

\subsection{Proof Sketch of Theorem~\ref{thm:stationary}}\label{sec:stationarysketch}
In this section, we describe the high-level proof idea of Theorem~\ref{thm:stationary}.
Note that Theorem~\ref{thm:stationary} characterizes the first order stationary points of $\tracehess$ on $\manifold$, i.e.,  
points $\thetastar$ on $\manifold$ for which 
$\nabla \tracehess(\thetastar) = 0$.
The starting point of the proof is the observation that the gradients $D f_i(\theta)$ form a basis for the normal space at $\theta$.
 \begin{lemma}[Basis for the normal space]\label{lem:normalbasis}
     For every $\theta$, the set of vectors $\brackets{Df_i(\theta)}_{i=1}^n$ form a basis for the normal space $\normalspace{\theta}$ of $\manifold$ at $\theta$.
 \end{lemma}

For a stationary point $\theta^*$ with $\nabla F(\theta^*) = 0$, the Euclidean gradient at $\thetastar$ should be in the normal space $\normalspace{\theta^*}$. But by Lemma~\ref{lem:normalbasis}, 
this means there exist coefficients $\brackets{\alpha_i}_{i=1}^n$ such that
\begin{align}
    D(\tracehess)(\thetastar) = \sum_{i=1}^n 2\alpha_i Df_i(\thetastar).\label{eq:linearcombination}
\end{align}
To further understand what condition~\eqref{eq:linearcombination} means for our two-layer network~\eqref{eq:basicmodel}, we first state a result from prior work in Lemma~\ref{lem:tracehessformula} (see e.g.\ \cite{blanc2020implicit}) regarding the trace of the Hessian of the mean squared loss $\loss$ for some parameter $\theta \in \manifold$ (see Section~\ref{sec:prooflem2} for a proof).
\begin{lemma}\label{lem:tracehessformula}
    For the loss on the dataset defined in Equation~\eqref{eq:meansquaredloss} and parameters $\theta$ with $\loss(\theta) = 0$, we have
    \begin{align*}
        \tracehess(\theta) = \sum_{i=1}^n \|D f_i(\theta)\|^2.
    \end{align*}
\end{lemma}
Using Lemma~\ref{lem:tracehessformula}, we can explicitly calculate the trace of Hessian for our two-layer network in \eqref{eq:basicmodel}. 
\begin{lemma}[Trace of Hessian in two-layer networks]\label{lem:tracehesstwolayer}
    For the neural network model~\eqref{eq:basicmodel} and the mean squared loss $\loss$ in Equation~\eqref{eq:meansquaredloss}, for any  $\theta$ with $\loss(\theta) = 0$ we have
    \begin{align}
        \tracehess(\theta) = \sum_{i=1}^n \sum_{j=1}^m \phi'(\theta_j^\top x_i)^2.\label{eq:traceofhessian}
    \end{align}
\end{lemma}

Now we are ready to prove Theorem~\ref{thm:stationary}.
\begin{proof}[Proof of Theorem~\ref{thm:stationary}]
From Equation~\eqref{eq:linearcombination}  and by explicitly calculating the gradients $\{Df_i(\thetastar)\}_{i=1}^n$ using Lemma~\ref{lem:tracehesstwolayer}, we have
\begin{align*}
    &\stack{\sum_{i=1}^n 2\phi'({\thetastarj}^\top x_i)\phi''({\thetastarj}^\top x_i) x_i}_{j=1}^m \\
    &= \sum_{i=1}^n 2\alpha_i \stack{\phi'({\thetastarj}^\top x_i) x_i}_{j=1}^m.
\end{align*}
But using our assumption that the data points $\brackets{x_i}_{i=1}^n$ are linearly independent, we have for all $i \in [n]$
 and $j \in [m]$, $\phi''({\thetastarj}^\top x_i) = \alpha_i$.
 Now, because $\phi'''$ is positive and $\phi''$ is strictly monotone, its inverse is well-defined:
 \begin{align}
     {\thetastarj}^\top x_i = \nu_i \triangleq \phi''^{-1}(\alpha_i),\label{eq:dotproducts}
 \end{align}
 This implies $\phi({\thetastarj}^\top x_i) = \phi(\nu_i)$. Therefore,
 \begin{align}
     y_i = \sum_{j=1}^m \phi({\thetastarj}^\top x_i) = m\phi(\nu_i).\label{eq:akhar}
 \end{align}
 But note that from strict monotonicity of $\phi$ from Assumption~\ref{assump:one}, we get that it is invertible. Therefore, Equation~\eqref{eq:akhar} implies
 \begin{align*}
     \nu_i = \phi^{-1}(y_i/m),
\quad \text{and} \quad 
     \alpha_i = \phi''(\phi^{-1}(y_i/m)).
 \end{align*}
 This characterizes the first order optimal points of $\tracehess$. 
 
 On the other hand, note that $\tracehess \geq 0$, so its infimum over $\manifold$ is well-defined. Moreover, from Lemma~\ref{lem:boundedregion} the set of $\theta$'s with $\tracehess(\theta) \leq c$ for arbitrary $c$ is bounded in $\mathbb R^{md}$. Therefore $\tracehess$ has a global minimizer on $\manifold$. 
 But the global minimizers are also stationary points and from Equation~\eqref{eq:dotproducts} we see that all of the first order optimal points have the same value of $\tracehess$. Therefore, all first order optimal points are global optima and satisfy Equation~\eqref{eq:dotproducts}. 
 \end{proof}

\paragraph{\khashared{Results for General Second Layer Weights:}} Note that if the weights of the second layer are fixed arbitrarily to $(w_j)_{j=1}^m$, then Equation~\eqref{eq:dotproducts} simply turns into ${\theta^*_j}^\top x_i = {\phi''}^{-1}(\alpha_i/w_j)$; namely, the $\phi''$ image of the feature matrix $\Theta^* X$ is rank one. Interestingly, this shows that in the general case, the right matrix to look for obtaining a low-rank structure is the embedding of the feature matrix by the second derivative of the activation (for the cube activation this embedding is the feature matrix itself.) 
Next, we investigate the convergence of the gradient flow in Equation~\eqref{eq:gradientflow}.

\subsection{Proof Sketch of Theorem~\ref{thm:rateofconvergenceone}}\label{sec:proofsketchthm3}

The first claim~\ref{claim:second} of Theorem~\ref{thm:rateofconvergenceone} is that after time $t\ge \tilde \Omega(\log(1/\epsilon))$, the gradient will always be smaller than $\epsilon$. It is actually trivial to prove that there exists some time at most $t = \tilde O(1/\epsilon^2)$ such that $\|\nabla \tracehess(\theta(t))\| \leq \epsilon$ without any assumption, based on a standard argument which is a continuous version of descent lemma. Namely, if the gradient of $\tracehess$ remains larger than some $\epsilon > 0$ along the flow until time $t >  \tracehess(\theta(0))/\epsilon^2$, then we get a contradiction:
\begin{align*}
    &\tracehess(\theta(0)) - \tracehess(\theta(t))\\
    &= \int_{0}^t \Big\|\nabla \tracehess(\theta(t))\big\|^2 > \tracehess(\theta(0)).
\end{align*}
Our novelty here is a new characterization of the loss landscape under Assumptions~\ref{assump:one},~\ref{assump:two} (Lemma~\ref{lem:psdness} proved in Appendix~\ref{sec:localgconv}).
\begin{lemma}[PSD Hessian when gradient vanishes]\label{lem:psdness}
    Suppose that activation $\phi$ satisfies Assumptions~\ref{assump:one},~\ref{assump:two}, and~\ref{assump:three}.
    Consider a point $\theta$ on the manifold where the gradient is small, namely $\|\nabla_\theta \trace\hess(\theta)\|\leq \sqrt \mu \beta$.
    Then, the Hessian of $\tracehess$ on the manifold is PSD at point $\theta$, or equivalently $\tracehess$ is locally $g$-convex on $\mathcal M$ around $\theta$.
\end{lemma}

Lemma~\ref{lem:psdness} implies that whenever the gradient is sufficiently small, the time derivative of the squared gradient norm will also be non-positive:
\begin{align*}
    &\frac{d}{dt}\|\nabla \tracehess(\theta(t))\|^2 \\
    &= -\nabla \tracehess(\theta(t))^\top \nabla^2 \tracehess(\theta(t)) \nabla \tracehess(\theta(t))\leq 0,
\end{align*}
Therefore, once the gradient is sufficiently small at some $t$, it will always remain small for all $t'>t$. In fact, we show that the Hessian of $\tracehess$ on the manifold is strictly positive with a uniform $\varrho_1 \varrho_2\mu$ lower bound in a suitable subspace which includes the gradient (Lemma~\ref{lem:hessstrongconvexity}). Then
\begin{align*}
       &\frac{d\|\nabla \tracehess(\theta(t))\|^2}{dt} \\
       &= -\nabla \tracehess(\theta(t))^\top \nabla^2 \tracehess(\theta(t))\nabla \tracehess(\theta(t)) \\
       &\leq -\varrho_1 \varrho_2\mu \|\nabla \tracehess(\theta(t))\|^2,
    \end{align*}
    which further implies a linear convergence of gradient norm:
    \begin{align*}
        \|\nabla \tracehess(\theta(t))\|^2 \leq \|\nabla \tracehess(\gamma(\theta(t_0)))\|^2 e^{-(t - t_0)\varrho_1\varrho_2 \mu}.
    \end{align*}

Next, we state the high level ideas that we use to prove Lemma~\ref{lem:psdness}, which is the key to proving Theorem~\ref{thm:rateofconvergenceone}. Before that, we review some necessary background.
\paragraph{Computing the Hessian on $\manifold$.}
To prove Lemma~\ref{lem:psdness}, we calculate the Hessian of $F \triangleq \tracehess$ on the manifold in Lemma~\ref{lem:hessianformula}. The Hessian of a function $F$ on $\Rr$ is defined as $D^2F[u,w] = \langle D(DF(\theta))[u], w\rangle$, where $DF(\theta)$ is the usual Euclidean gradient of $F$ at $\theta$, and $D(.)[u]$ denotes directional derivative. To calculate the Hessian on the manifold, one needs to substitute the Euclidean gradient $DF(\theta)$ by the gradient $\nabla F(\theta)$ on the manifold. Moreover, $D(DF(\theta))[u]$ has to be substituted by the covariant derivative $\nabla_u \nabla F(\theta)$, a different differential operator than the usual derivative. 
We recall the characterization of the covariant derivative as the projection of the Euclidean directional derivative onto the tangent space. For more background on covariant differentiation, we refer the reader to Appendix~\ref{sec:rgeometry}.
\begin{fact}\label{fact:covariant}
    For vector fields $V,W$ on $\hypersurface$, we have $\nabla_V W(\theta) = \thetaproj (DW(\theta))[V]$.
\end{fact}
 We also recall the definition of Hessian $\nabla^2 F$ on $\hypersurface$ based on covariant derivative.
\begin{fact}~\label{fact:hessian}
    The Hessian of $F$ at point $\theta$ on $\hypersurface$ is given by
    $\nabla^2 F(w,u) = \langle \nabla_w \nabla F, u\rangle$.
\end{fact}

Next, we state Lemma~\ref{lem:hessianformula}, which is proved in~\Cref{sec:proofhessianformula2}.

\begin{lemma}[Hessian of the implicit regularizer on the manifold]\label{lem:hessianformula}
Recall that $\brackets{Df_i(\theta)}_{i=1}^n$ is a basis for the normal space $\normalspace{\theta}$ according to Lemma~\ref{lem:normalbasis}. Let $\alpha' = (\alpha'_i)_{i=1}^n$ be the coefficients representing $\nproj(D(\tracehess)(\theta))\in \normalspace{\theta}$ in the basis $\brackets{Df_i(\theta)}_{i=1}^n$, i.e.
\begin{align*}
    \nproj\Big(D(\trace\hess)(\theta)\Big) = \sum\nolimits_{i=1}^n \alpha'_i Df_i(\theta).
\end{align*}
Then, the Hessian of $\tracehess$ on $\manifold$ can be explicitly written (in the Euclidean chart $\Rr$) using $\alpha'$ as
    \begin{align}
        &\nabla^2 \tracehess(\theta)[u,w] \notag\\
        &= D^2 \tracehess(\theta)[u,w] - \sum\nolimits_{i=1}^n \alpha'_i D^2 f_i(\theta)[u,w],\label{eq:hessformula}
    \end{align}
    for arbitrary $u,w\in \tangentspace{\theta}$, where $D^2$ and $\nabla^2$ denote the normal Euclidean Hessian and Hessian over the manifold.
\end{lemma}
Observe in the formula of $\nabla^2(\tracehess)$ in Equation~\eqref{eq:hessformula} the first term is just the normal Euclidean Hessian of $\tracehess$ while we get the second ``projection term'' due to the difference of covariant differentiation with the usual derivative.
Finally, to prove the second claim~\ref{claim:third} of \Cref{thm:rateofconvergenceone}, we prove \Cref{lem:closetoopt} in~\Cref{sec:proofclosetoopt} which shows that the small gradient norm of $\|\nabla \tracehess(\theta)\|\leq \delta$ implies the (approximate) alignment of features.
\begin{lemma}[Small gradient implies close to optimum]\label{lem:closetoopt}
    Suppose $\|\nabla \tracehess(\theta)\| \leq \delta$. Then, for all $i\in [n], j\in [m]$,
        $\Big|\theta_j^\top x_i - {\theta^*_j}^\top x_i\Big| \leq \delta/(\sqrt \mu \varrho_1 \varrho_2),$
    for $\theta^*$ in Equation~\eqref{eq:effectofdepth}.
\end{lemma}

\section{Experimental Setup}
We examine our theory on the convergence to a rank one feature matrix in Figure~\ref{fig:one} via a synthetic experiment by considering a network with $m = 10$ neurons on ambient dimension $d = 3$ and $n = 3$ data points. We further pick learning rate $\eta = 0.05$ and noise variance $\sigma = 0.03$ for implementing label noise SGD. Each entry of the data points is generated uniformly on $[0,1]$, which is the same data generating process in all the experiments. As Figure~\ref{fig:one} shows, the second and third eigenvalues converge to zero which is predicted by Theorem~\ref{thm:stationary}. On the other hand, in a similar setting with the same learning rate $\eta = 0.05$ but larger $\sigma = 0.2$ we run a synthetic experiment with $d = 5$ this time smaller than the number of data points $n = 6$.  In this case, the feature matrix cannot be rank one anymore. However, one might guess that it still converges to something low-rank even though our theoretical result is shown only for the high dimensional case; in this case, the smallest possible rank to fit the data is indeed two, and surprisingly we see in Figure~\ref{fig:three} that except the first and second eigenvalues of the feature matrix, the rest go to zero. To further test extrapolating our theory to the low dimensional case ($d < n$), this time we use a higher number of data points $n=15$ with ambient dimension $d = 10$ and $30$ neurons, with learning rate $\eta = 0,1$ and noise variance $\sigma = 0.2$. We observe a very interesting property: even though the feature matrix is high rank in this case, the neurons tend to converge to one another, especially when they are close, as if there is a hidden attraction force between them. To visualize our experimental discovery, in Figure~\ref{fig:two} we plot the first two principal components (with the largest singular values) of the feature vectors for each neuron, and observe how it changes while running label noise SGD. We further pick $12$ neurons for which the first and second components are bounded in $[0,0.1]$ and $[0,0.5]$, respectively. As one can see in Figure~\ref{fig:two}, most of the principal components of the neurons except possibly one outlier are converging to the same point. This is indeed the case for the feature vectors of neurons as well, namely, they collapse into clusters, but unfortunately, we only have two dimensions to visualize this, and we have picked the most two effective indicators (the first and second principal components.)

 \section{Conclusion}\label{sec:conclusion}
In this paper, we take an important step toward understanding the implicit bias 
of label noise SGD with trace of Hessian induced regularizer for a class of two layer networks. We discover an intriguing relationship between this induced regularizer and the low-rank simplicity bias conjecture of neural networks proposed in~\cite{huh2021low}: we show that by initializing the neurons in the subspace of high-dimensional input data, all of the neurons converge into a single vector. Consequently, for the final model, (i) the rank of the feature  matrix is effectively one, and (ii) the functional representation of the final model is \emph{very simple}; specifically, its sub-level sets are half-spaces. To prove this, in spite of the lack of convexity, we uncover a novel structure in the landscape of the loss regularizer: the trace of Hessian regularizer becomes locally geodetically convex at points that are approximately stationary. Furthermore, in the limit of step size going to zero, we prove that  label noise SGD or 1-SAM converge \emph{exponentially fast} to a global minimizer. 

Furthermore, generalizing the class of activations that enjoy fast convergence or proving the existence of fundamental barriers and handling the case of low-dimensional input data are interesting future directions. 
Based on the compelling empirical evidence from \cite{huh2021low} and our results for two-layer networks, we hypothesize that a low-rank simplicity bias can also be shown for deeper networks, possibly using the sharpness minimization framework.

\section*{Impact Statement}


This paper presents work whose goal is to advance the field of Machine Learning. There are many potential societal consequences of our work, none which we feel must be specifically highlighted here.


\section*{Acknowledgement}
This research was supported in part by
Office of Naval Research grant N00014-20-1-2023 (MURI ML-SCOPE), NSF AI Institute TILOS  (NSF CCF-2112665), NSF award  2134108, the Alexander von Humboldt Foundation, and the NSF TRIPODS program (award DMS-2022448).

\newpage

\bibliography{arxiv-new/template}
\bibliographystyle{plainnat}
\newpage
\appendix
\onecolumn
\section{Proof of Theorem~\ref{thm:main_intro}}\label{sec:high-level-app}
     \Cref{thm:main_intro} is a direct combination of Theorem 4.6 of \citet{li2021happens}, \Cref{thm:stationary}, and \Cref{thm:rateofconvergenceone}.
    For the convenience of the reader, we restate Theorem 4.6 of~\citet{li2021happens} here. To avoid any confusions, we have specialized Theorem 4.6 of~\citet{li2021happens} into our case for label noise SGD with our notation. We remind the reader that we use $\theta(t)$ for the gradient flow at time $t$ and $\theta_t$ for the $t$th iteration of label noise SGD (Note that we do not necessarily set $\theta(0)$ equal to $\theta_0$, see the theorem below). Before restating Theorem 4.6, we first restate its two prerequisite assumptions.
    \begin{assumption}[Assumption 3.1. in~\citet{li2021happens}]\label{assumptionlione}
        Assume that the loss $L:\mathbb R^d \rightarrow \mathbb R$ is $\mathcal C^3$ function, and that $\manifold$ is a $(md - n)$-dimensional $\mathcal C^2$-submanifold of $\mathbb R^{md}$, where for all $\theta \in \manifold$, $\theta$ is a local minimizer of $\loss$ and $\text{rank}(\nabla^2 \loss(\theta)) = n$. 
    \end{assumption}
    \begin{assumption}[Assumption 3.2. in~\citet{li2021happens}]\label{assumptionlitwo}
        For arbitrary $\theta \in \mathbb R^{md}$, let $\phii(\theta)\in \manifold$ be the point of convergence of the gradient flow according to $\loss$ starting from $\theta$, assuming it exists. Assume that $U$ is an open neighborhood of $\manifold$ satisfying that gradient flow of $\loss$ starting in $U$ converges to some point in $\manifold$, i.e., $\forall \theta \in U$, $\phii(x) \in \manifold$.
    \end{assumption}
    \begin{theorem}[Theorem 4.6 in~\citet{li2021happens}]\label{thm:lirestatement}
        Suppose the loss function $\loss$, the manifold of local minimizer $\manifold$ and the open neighborhood $U$ satisfy Assumptions 3.1 and 3.2 stated above, and $\theta_0 \in U$ is fixed for all choices of step size $\eta > 0$. If the gradient flow of trace of hessian~\eqref{eq:gradientflow} has a global solution $\theta(t)$ with $\theta(0) = \phii(\theta_0)$ and $\theta(t)$ never leaves $U$, i.e. $\mathbb P(\theta(t)\in U, \forall t\geq 0) = 1$. Then for any $T > 0$, $\theta_{\lfloor T/(\eta^2 \sigma^2)\rfloor}$ converges in distribution to $\theta(T)$ as $\eta \rightarrow 0$.
    \end{theorem}

    First note that from Lemma~\ref{lem:firtflowconverge}, the gradient flow in the limit of step size going to zero always converges to the manifold of zero, independent of the initialization. 
    Therefore, we can set the neighborhood $U$ in Theorem 4.6 of~\cite{li2021happens} to be the whole $\Rr$.
    Now we need to check Assumptions 3.1. and 3.2. in~\cite{li2021happens}.

    \textbf{Checking Assumption~\ref{assumptionlione}} 
    The smoothness of the squared loss $\loss$ defined in Equation~\eqref{eq:meansquaredloss} follows from the smoothness of the activation function $\phi$. Furthermore, as we derived in Equation~\eqref{eq:hessianformulaa}, we have $\nabla^2 \loss = 2\sum_{i=1}^n Df_i(\theta)Df_i(\theta)^\top$, and from the non-singularity of the Jacobian of $f$ stated in Lemma~\ref{lem:nonsingular} and the linear independence of the inputs $x_i$'s due to Assumption~\eqref{assump:two}, we get that $Df_i$'s are always linearly independent, and as a result, the Hessian $\nabla^2 \loss$ is always rank $n$. Combining these with the fact that $\manifold$ is a well-defined smooth submanifold of $\mathbb R^{md}$ due to Lemma~\ref{lem:manifoldwelldefined} fully checks Assumption~\ref{assumptionlione}.

    \textbf{Checking Assumption~\ref{assumptionlitwo}}
    We take $U$ to be the entire $\mathbb R^{md}$. Then according to Lemma~\ref{lem:firtflowconverge}, starting from any $\theta_0$, the gradient flow of the loss converges to a point on the manifold $\manifold$.

    Since Assumptions~\ref{assumptionlione} and~\ref{assumptionlitwo} are satisfied, we can apply Theorem~\ref{thm:lirestatement}: namely, Theorem~\ref{thm:lirestatement} implies there exists a threshold $\eta_0$ such that for step size $\eta \leq \eta_0$, we have $\theta_{\lceil t/\eta^2 \rceil}$ is $\delta$ to $\theta(t)$ in distribution, where recall that $\theta(t)$ is
    the Riemannian gradient flow according to the trace of Hessian defined in Equation~\eqref{eq:gradientflow}, initialized at $\phii(\theta_0)$. 

    In order to obtain Lipschitz bounds for $\theta(t)$ which evolves according to~\eqref{eq:gradientflow}, we need to obtain information about the initial point $\theta(0)$ on the manifold, which is the limiting point $\tilde \theta$ of the trace of hessian gradient flow. This limiting point is defined in Lemma~\ref{lem:firtflowconverge}. 
    In particular, in that Lemma we show that the trace of Hessian of the final point is bounded as
    \begin{align*}
        \tracehess(\theta(0)) = \tracehess (\tilde \theta) \leq 2\tracehess (\theta_0) + \frac{2}{\sqrt C}\sqrt{\loss(\theta_0)},
    \end{align*}
    for $C = 4m\mu \varrho_1^2$. On the other hand, note that from Theorem~\ref{thm:rateofconvergenceone}, $\theta(t)$ converges to $\tilde \epsilon$-proximity of a global minimizer $\theta^*$ of trace of Hessian on $\manifold$ in the sense of Equation~\eqref{eq:dotproductproximity} after time at most 
    $$\frac{\tracehess(\theta(0))}{\mu \beta^2} + \frac{\log( \beta^2/(\varrho_1^2\varrho_2^2{\tilde \epsilon}^2) \vee 1)}{\varrho_1\varrho_2 \mu}
    \leq 
    \frac{2\tracehess (\theta_0) + \frac{2}{\sqrt C}\sqrt{\loss(\theta_0)}}{\mu \beta^2} + \frac{\log( \beta^2/(\varrho_1^2\varrho_2^2{\tilde \epsilon}^2) \vee 1)}{\varrho_1\varrho_2 \mu} \coloneqq \tilde T.$$

    
    Now using Theorem~\ref{thm:lirestatement} for $T = \tilde T$, we get that $\theta_{\lfloor\tilde T/(\eta^2 \sigma^2)\rfloor}$ converges to $\theta(\tilde T)$ in distribution. But $\theta(\tilde T)$ is a deterministic point mass, so convergence in distribution implies convergence in probability. Namely, for every $\tilde \epsilon>0$, there exists a threshold $\eta_0$ such that for step size $\eta \leq \eta_0$,
    with probability $1-\epsilon$, 
    \begin{align}
        \big|{\theta(\tilde T)}_j^\top x_i - {\theta_{\lfloor\tilde T/(\eta^2 \sigma^2) \rfloor}}_j^\top x_i \big| \leq \tilde \epsilon.\label{eq:guaranteee}
    \end{align}
    
    On the other hand, from Theorem~\ref{thm:rateofconvergenceone}, after 
    $\lfloor\tilde T/(\eta^2\sigma^2)\rfloor$ iterations of label noise SGD, with probability at least $1-\epsilon$, ${{\theta_{\tilde T}}_j}^\top x_i$ is close to ${\theta^*_j}^\top x_i$ for all $j\in [m]$ and $i \in [n]$:
    \begin{align}
        \big|{\theta}(\tilde T)_j^\top x_i - {\theta^*}_j^\top x_i\big| \leq \tilde \epsilon.\label{eq:guaranteeee}
    \end{align}
    Combining Equations~\eqref{eq:guaranteeee} and~\eqref{eq:guaranteee}:
    \begin{align}
        \big|{\theta^*}_j^\top x_i - {\theta_{\lfloor\tilde T/(\eta^2 \sigma^2) \rfloor}}_j^\top x_i \big| \leq 2\tilde \epsilon.\label{eq:hasel}
    \end{align}
    Moreover, note that from Lemma~\ref{lem:boundedregion}, there exists $\delta',\epsilon'$ depending on the activation $\phi$ such that for all $i\in [n], j\in [m]$ and times $t\geq 0$: 
    \begin{align*}
        \big|\theta(t)_j^\top x_i - z^*\big| \leq \tracehess (\theta(0))/(\epsilon' \delta') + \epsilon'/2
        \leq \Big(2\tracehess(\theta_0) + \frac{2}{\sqrt C}\sqrt{\loss(\theta_0)}\Big)/(\epsilon'\delta') + \epsilon'/2 \coloneqq R,
    \end{align*}
    where $z^*$ is the global minimizer of $\phi'$ and we defined the RHS to be parameter $R$.
    This means that $\theta(t)_j^\top x_i$ always remains within distance $R$ from $z^*$ along the gradient flow on the manifold. Combining this with Equation~\eqref{eq:guaranteee} implies that the points ${\theta}(\tilde T)_j^\top x_i$, ${\theta_{\lfloor\tilde T/(\eta^2 \sigma^2) \rfloor}}_j^\top x_i$, and ${\theta^*}_j^\top x_i$ are within distance at most $R + \tilde \epsilon$ of $z^*$. On the other hand, the functions $\tracehess$ and $\loss$ are smooth, and in particular have bounded Lipschitz constants $L_1,L_2$ in the hypercube $[-(R+\tilde \epsilon), (R+\tilde \epsilon)]^{mn}$. Using these Lipschitz constants with Equation~\eqref{eq:hasel} implies
    
    \begin{align*}
        \Big|\tracehess(\theta_{\lfloor\tilde T/\eta^2\rfloor}) - \tracehess(\theta^*)\Big| \leq 2L_1\tilde \epsilon,\\
        \Big| \loss(\theta_{\lfloor\tilde T/\eta^2\rfloor}) - \loss(\theta^*)\Big| \leq 2L_2\tilde \epsilon. 
    \end{align*}
    Picking $\tilde \epsilon \leq \min(\epsilon/(2L_1), \epsilon/(2L_2))$ completes the proof.

\begin{figure}[ht]
\vskip 0.2in
\begin{center}
\centerline{\includegraphics[width=350pt]{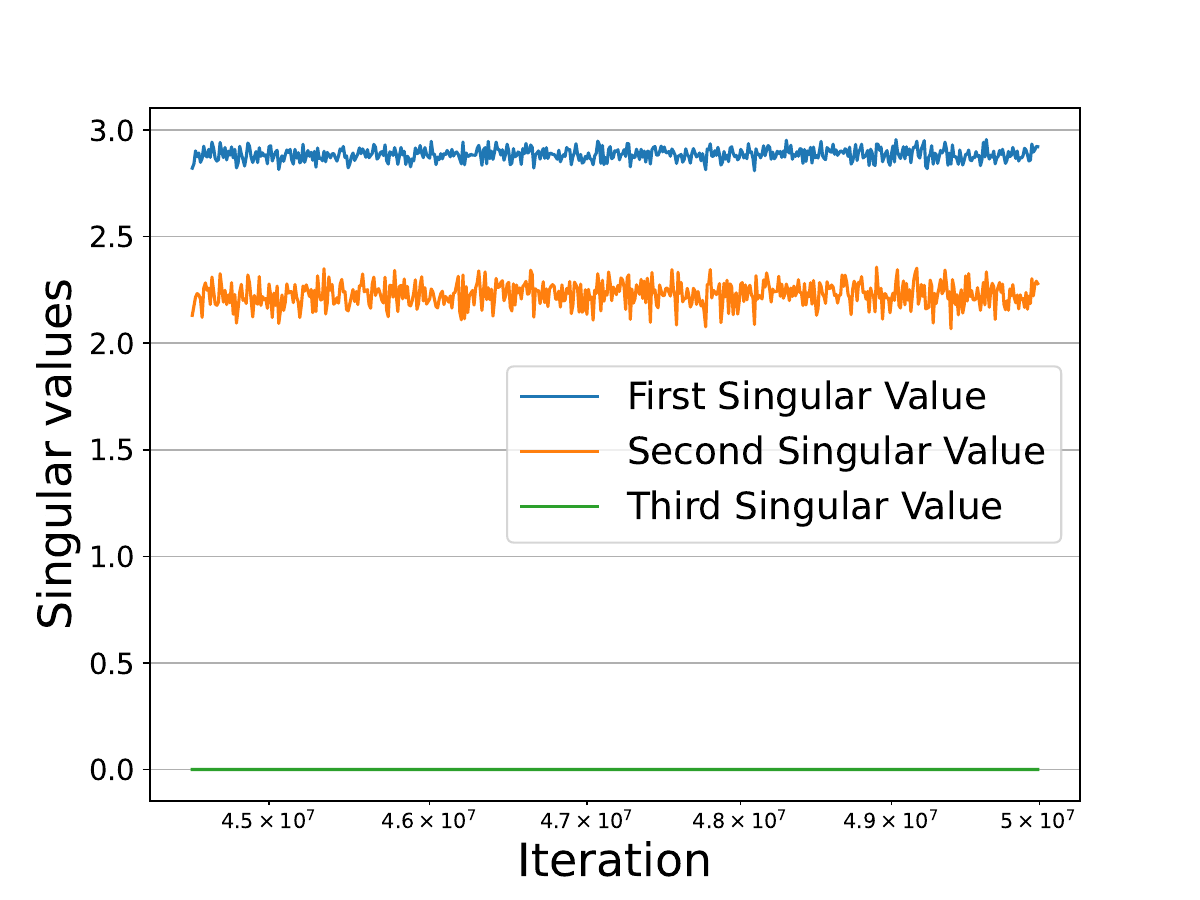}}
\caption{The feature embeddings of neurons collapse into clusters in the low dimensional regime $d < n$. The plot visualizes this phenomenon using the first and second principal components of the feature embedding $\Theta^* X$.}
\label{fig:three}
\end{center}
\vskip -0.2in
\end{figure}

\section{Full proof of Theorem~\ref{thm:rateofconvergenceone}}\label{sec:convergence-app}

In order to prove Theorem~\ref{thm:rateofconvergenceone}, we take these steps:
\begin{itemize}
    \item In Section~\ref{sec:localgconv}, we show that the norm of the gradient of $\tracehess$ becomes decreasing at approximate stationary points in Lemma~\ref{lem:gradientdecay0}. To this end, we show that under Assumptions~\ref{assump:one},~\ref{assump:two},~\ref{assump:three}, at each point $\theta \in \manifold$, we either have a large enough gradient or we are in a neighborhood of the global optima of the trace of Hessian in which trace of Hessian becomes g-convex as stated in Lemma~\ref{lem:psdness}.
    \item In Section~\ref{sec:slocalgconv}, Lemma~\ref{lem:gradientdecay}, we show that the norm of the gradient of $\tracehess$ decays exponentially fast under Assumptions~\ref{assump:one},~\ref{assump:two}, and~\ref{assump:three}. To show this, we prove Lemma~\ref{lem:hessstrongconvexity} in which we show under Assumptions~\ref{assump:one},~\ref{assump:two},and~\ref{assump:three} the Hessian of $\tracehess$ on $\manifold$ becomes positive definite at approximate stationary points.
     \item Finally, we combine Lemmas~\ref{lem:gradientdecay0},~\ref{lem:gradientdecay}, and~\ref{lem:closetoopt} to show Theorem~\ref{thm:rateofconvergenceone} in Appendix~\ref{sec:proofthm2}.
\end{itemize}

\subsection{Local g-convexity at approximate stationary points}\label{sec:localgconv}
In this section we prove Lemma~\ref{lem:gradientdecay0}.

\begin{lemma}[PSD Hessian when gradient vanishes]\label{lem:gradientdecay0}
The norm of the gradient $\|\nabla \tracehess(\theta(t))\|$ on the gradient flow~\eqref{eq:gradientflow} becomes decreasing with respect to $t$ at $\sqrt{\mu}\beta$-approximate stationary points, i.e. when $\|\nabla \tracehess(\theta(t))\| < \sqrt \mu \beta$.
\end{lemma}


To prove Lemma~\ref{lem:gradientdecay0}, we prove Lemma~\ref{lem:psdness} in the rest of this section.
Note that from the continuity of $\big\|\nabla \tracehess(\theta)\big\|$, we see that the Hessian of the implicit regularizer $\tracehess$ is PSD in a neighborhood of a $\sqrt \mu \beta/2$-stationary point $\theta \in \manifold$. This implies that $\tracehess$ is g-convex in a neighborhood of such $\theta$. We state a short proof of this in Lemma~\ref{lem:localgconvexity} in the appendix. 
Even though under Assumption~\ref{assump:one} we get that $\tracehess$ has a convex extension in $\Rr$, it is not necessarily g-convex over the manifold. This is because the Hessian over the manifold is calculated based on covariant differentiation, which is different from the normal Euclidean derivative. In particular, for a hypersurface $\manifold$ in $\Rr$, the covaraint derivative is obtained by taking the normal Euclidean derivative from the tangent of a curve on $\manifold$, then projecting the resulting vector back onto the tangent space. This projecting back operation in the definition of Covariant derivative results in an additional ``projection term" in the algebraic formula of the Hessian of $\tracehess$ on $\manifold$, in addition to its normal Euclidean Hessian. The inversion in the projection operator makes it fundamentally difficult to derive good estimates on the projection term that are relatable to the Euclidean Hessian of $\tracehess$. Nonetheless, we discover a novel structure in $\manifold$ which is that even though 
 we cannot control the projection term in the Hessian of $\tracehess$ globally, but for $\epsilon$-stationary points $\theta$ on the manifold, we can relate it to the approximate stationarity condition (equivalently approximate KKT condition) at $\theta$. This relation enables us to control the projection term and show that the Hessian of $\tracehess$ on $\manifold$ is indeed PSD at approximate stationary points. 
To prove Lemma~\ref{lem:psdness}, first we calculate the formula for the Hessian of an arbitrary function $F$ on a hypersurface $\manifold$ embedded in $\Rr$, in Lemma~\ref{lem:hessianformulalowlevel}. We then use this formula to calculate the Hessian of $\tracehess$ on $\manifold$ in Lemma~\ref{lem:hessianformula}. Finally we use this formula to prove Lemma~\ref{lem:psdness} in Section~\ref{sec:proofsketchthm3}.

Now we derive a general formula for the Hessian of any smooth function $F$ on the manifold in Lemma~\ref{lem:hessianformulalowlevel}, proved in Appendix~\ref{sec:prooflem5}.
\begin{lemma}[Hessian formula on a submanifold of $\Rr$]\label{lem:hessianformulalowlevel}
For any smooth function $F$ defined on $\Rr$, the Hessian of $F$ on $\manifold$ is given by:
\begin{align*}
    \nabla^2 F(\theta)[u,w] = D^2 F[u,w] - D^2f(\theta)[u,w]^\top  (Df(\theta) Df(\theta)^\top)^{-1} Df(\theta)(DF(\theta)),
\end{align*}
for $u,w\in \tangentspace{\theta}$.
\end{lemma}

Next,
we prove Lemma~\ref{lem:hessianformula}, 
exploiting the formula that we derived in Lemma~\ref{lem:hessianformulalowlevel} for the Hessian of a general smooth function $F$ over $\manifold$.

\subsubsection{Proof of Lemma~\ref{lem:hessianformula}}
\begin{proof}[Proof]\label{sec:proofhessianformula2}
Using the formula for the Hessian of a function over the manifold in Lemma~\ref{lem:hessianformulalowlevel}, we have
    \begin{align*}
    &\nabla^2 \trace\hess(\theta)[u,w] \\
    &= D^2 \tracehess(\theta)[u,w] - D^2 f(\theta)[u,w]^\top (Df(\theta) {Df(\theta)}^\top)^{-1}Df(\theta)(D\trace\hess(\theta))\\
    &= D^2 \tracehess(\theta)[u,w] \\
    &- D^2 f(\theta)[u,w]^\top (Df(\theta) {Df(\theta)}^\top)^{-1}Df(\theta)\Big(\nabla_\theta \trace D^2 L(\theta) + 2\sum_{i=1}^n \alpha'_i Df_i(\theta)\Big)\\
    & \stackrel{\eqref{eq:mylabeltwo}}{=} D^2 \tracehess(\theta)[u,w] 
    - 2 D^2 f(\theta)[u,w]^\top (Df(\theta) {Df(\theta)}^\top)^{-1}Df(\theta)\Big(\sum_{i=1}^n \alpha'_i Df_i(\theta)\Big)\numberthis\label{eq:mylabeltwo}\\
    & = D^2 \trace\hess[u,w] 
    - 2 D^2 f(\theta)[u,w]^\top (Df(\theta) {Df(\theta)}^\top)^{-1}(Df(\theta)Df(\theta)^\top)\alpha'\\
    & = D^2 \tracehess(\theta)[u,w] - 2\sum_{i=1}^n \alpha'_i D^2 f_i(\theta)[u,w], 
\end{align*}
where equality~\eqref{eq:mylabeltwo} follows from the fact that $\nabla_\theta(\tracehess) \in \tangentspace{\theta}$, which means it is orthogonal to $Df_i(\theta)$ for all $i\in [n]$.    
\end{proof}

Using Lemma~\ref{lem:hessianformula}, we prove Lemma~\ref{lem:psdness}. 
We restate Lemma~\ref{lem:psdness} here for the convenience of the reader.

\begin{lemma*}[see Lemma~\ref{lem:psdness}]
    Suppose that activation $\phi$ satisfies Assumptions~\ref{assump:one},~\ref{assump:two},~\ref{assump:three}.
    Consider a point $\theta$ on the manifold where the gradient is small, namely $\|\nabla_\theta \trace\hess(\theta)\|\leq \sqrt \mu \beta$.
    Then, the Hessian of $\tracehess$ on the manifold is PSD at point $\theta$, or equivalently $\tracehess$ is locally $g$-convex on $\mathcal M$ around $\theta$.
\end{lemma*}
\subsubsection{Proof of Lemma~\ref{lem:psdness}}
\begin{proof}[Proof]
    Suppose $\|\nabla_\theta \tracehess(\theta)\| = \delta$.
    Write the projection of the Euclidean gradient $D(\tracehess)(\theta)$ onto the normal space $\normalspace{\theta}$ in the row space of $Df(\theta)$, i.e. there exists a vector $\alpha' = (\alpha_i')_{i=1}^n$ such that 
    \begin{align*}
         \nproj[D(\tracehess)(\theta)] \triangleq 2\sum_{i=1}^n \alpha'_i Df_i(\theta),
    \end{align*}
 where $P^N_\theta$ is the projection onto the normal space at point $\theta$ on the manifold. Then, if we decompose $D(\tracehess)(\theta)$ into the part in the tangent space and the part in the normal space, because the part on the tangent space is the gradient of $\tracehess$ on the manifold, we have
\begin{align*}
    D(\trace\hess)(\theta) = \nabla_\theta \trace D^2 L(\theta) + 2\sum_{i=1}^n \alpha'_i Df_i(\theta).
\end{align*}
Now from the assumption that the norm of $\nabla \trace\hess(\theta)$ is at most $\delta$ and using the formula for $\tracehess$ in Lemma~\ref{lem:tracehesstwolayer}, we have
\begin{align*}
    \delta^2 = \Big\|\nabla_\theta \trace\hess(\theta)\Big\|^2 &= \Big\|D(\tracehess)(\theta) - 2\sum_{i=1}^n \alpha_i' \stack{\phiprime x_i}_{j=1}^m\Big\|^2\\
    &=4\sum_{j=1}^m \Big\|\sum_{i=1}^n (\phitwoprime - \alpha_i')\phiprime x_i\Big\|^2,
\end{align*}
which implies
\begin{align*}
    4\mu \sum_{i=1}^n (\phitwoprime - \alpha_i')^2 \phiprime^2 \leq 4\Big\|\sum_{i=1}^n (\phitwoprime - \alpha_i')\phiprime x_i\Big\|^2 \leq \delta^2.
\end{align*}
This further implies $\forall i\in [n], j \in [m]$:
\begin{align}
    |(\phitwoprime - \alpha_i') \phiprime| \leq \delta/(2\sqrt{\mu}).\label{eq:alphaiprimeupperbound}
\end{align}
Now using the $\beta$-normality of $\phi$, the fact that $\delta/\sqrt \mu \leq \beta$, and the positivity of $\phi'$ and $\phi'''$, we have for all $i \in [n]$ and $j \in [m]$, 
\begin{align}
2\Big|\phi''(\theta_j^\top x_i)(\alpha_i' - \phi''(\theta_j^\top x_i))\Big| \leq \phi'(\theta_j^\top x_i)\phi'''(\theta_j^\top x_i).\label{eq:giventhat}    
\end{align}
On the other hand, using Lemma~\ref{lem:hessianformula}, we can explicitly calculate the Hessian of $\tracehess$ as
\begin{align*}
    &\nabla^2 \trace\hess(\theta)[u,u] = D^2 \tracehess(\theta)[u,u] - 2\sum_{i=1}^n \alpha'_i D^2 f_i(\theta)[u,u]. 
\end{align*}
Now using the formula for the Euclidean Hessian of $\tracehess$ and $f_i$'s in   Lemmas~\ref{lem:derivativeequations} and~\ref{lem:derivativeequation2}, we get 
\begin{align*}
    \nabla^2 \trace\hess(\theta)[u,u]
    & = \sum_{j=1}^m \sum_{i=1}^n \Big(2\phiprime \phithreeprime + 2\phitwoprime^2\Big) (x_i^\top u_j)^2\\ 
    &\quad - \sum_{j=1}^m \sum_{i=1}^n 2\alpha'_i \phitwoprime (x_i^\top u_j)^2\\
    & = \sum_{j=1}^m \sum_{i=1}^n \Big(2\phiprime \phithreeprime + 2\phitwoprime(\phitwoprime - \alpha'_i)\Big) (x_i^\top u_j)^2\\ 
    &\geq \sum_{j=1}^m \sum_{i=1}^n \phiprime \phithreeprime (x_i^\top u_j)^2 \geq 0,\numberthis\label{eq:hessianbound}
\end{align*}
where the second to last inequality follows from Equation~\eqref{eq:giventhat}. Hence, the Hessian over the manifold is PSD at $\sqrt \mu \beta$-stationary points. 
\end{proof}

Based on Lemma~\ref{lem:psdness}, the proof of Lemma~\ref{lem:gradientdecay0} is immediate.
\subsubsection{Proof of Lemma~\ref{lem:gradientdecay0}}
\begin{proof}[Proof]
    To prove Lemma~\ref{lem:gradientdecay0} simply note that the derivative of the norm squared of the gradient becomes negative:
    \begin{align*}
        \frac{d}{dt}\|\nabla \tracehess(\theta(t))\|^2 = -\nabla \tracehess(\theta(t))^\top \nabla^2 \tracehess(\theta(t)) \nabla \tracehess(\theta(t))\leq 0.
    \end{align*}
\end{proof}
Next, we show a strict positive definite property for the hessian of approximate stationary points under our Assumptions~\ref{assump:one},~\ref{assump:two}, and~\ref{assump:three}.
\subsection{Strong local g-convexity}\label{sec:slocalgconv}
It turns out we can obtain an exponential decay on the norm of the gradient at approximate stationary points, given explicit positive lower bounds on $\phi'$ and $\phi'''$ stated in Assumption~\ref{assump:one}. As we mentioned, from Lemma~\ref{lem:boundedregion} the gradient flow does not exit a bounded region around $\thetastar$ on $\manifold$, which means Assumption~\ref{assump:one} automatically holds for some positive constants $\varrho_1, \varrho_2$ for the flow in Equation~\eqref{eq:gradientflow}, given that we have the weaker condition $\phi', \phi''' > 0$ everywhere; this is because of compactnesss of bounded regions in the Euclidean space.

We state the exponential decay in Lemma~\ref{lem:gradientdecay}. 

\begin{lemma}[Decay of norm of gradient]\label{lem:gradientdecay}
   Under assumptions~\ref{assump:one},~\ref{assump:one}, and~\ref{assump:two}, given time $t_0$ such that $\|\nabla \tracehess(\theta(t_0))\| \leq \sqrt \mu \beta$, then the norm of gradient of $\tracehess$ is exponentially decreasing for times $t \geq t_0$. Namely,
    \[
     \|\nabla \tracehess(\theta(t))\|^2 \leq \|\nabla \tracehess(\theta(t_0))\|^2 e^{-(t - t_0)\varrho_1\varrho_2 \mu}.
    \]
\end{lemma}

The key idea to obtain this exponential decay in Lemma~\ref{lem:gradientdecay} is that at approximate stationary points $\theta$, under Assumption~\ref{assump:one}, the Hessian of $\tracehess$ becomes strictly positive definite on a subspace of $\tangentspace{\theta}$ which is spanned by $\brackets{x_i}_{i=1}^n$ in the coordinates corresponding to $\theta_j$. Fortunately, the gradient $\nabla \tracehess(\theta)$ is in this subspace, hence we can obtain the following Lemma. We show this formally in Lemma~\ref{lem:hessstrongconvexity}, proved in Appendix~\ref{sec:proofhessstrongconvexity}.

\begin{lemma}[Strong convexity]\label{lem:hessstrongconvexity}
    Under Assumptions~\ref{assump:one},~\ref{assump:one}, and~\ref{assump:two} we have for $\sqrt \mu \beta$-stationary $\theta$, i.e. when $\|\nabla_\theta \tracehess\|\leq \sqrt \mu \beta$, 
    \begin{align*}
        {\nabla \tracehess(\theta)}^\top \Big(\nabla^2 \tracehess(\theta)\Big) \nabla \tracehess(\theta) \geq \varrho_1 \varrho_2\mu\|\nabla \tracehess(\theta)\|^2.
    \end{align*}
\end{lemma}

\subsubsection{Proof of Lemma~\ref{lem:gradientdecay}}
\begin{proof}[Proof]\label{sec:proofgradientdecay}

    Using Lemma~\ref{lem:hessstrongconvexity}, we get positive definite property of the Hessian $\nabla^2 \tracehess$ on the manifold by Lemma~\ref{lem:hessstrongconvexity}, which implies
    \begin{align*}
       \frac{d\|\nabla \tracehess(\theta(t))\|^2}{dt} &= -\nabla \tracehess(\theta(t))^\top \nabla^2 \tracehess(\theta(t))\nabla g(\theta(t)) \\
       & \leq -\varrho_1 \varrho_2\mu \|\nabla \tracehess(\theta(t))\|^2.
    \end{align*}
    This inequality implies an exponential decay in the size of the gradient, namely for any $t \geq t_0$
    \begin{align*}
        \|\nabla \tracehess(\theta(t))\|^2 \leq \|\nabla \tracehess(\theta(t_0))\|^2 e^{-(t - t_0)\varrho_1\varrho_2 \mu}.
    \end{align*}
\end{proof}

Next, we prove Theorem~\ref{thm:rateofconvergenceone}.
\subsection{Proof of Theorem~\ref{thm:rateofconvergenceone}}\label{sec:proofthm2}
    Using the non-decreasing property and exponential decay of the gradient at approximate stationary points that we showed in Lemmas~\ref{lem:gradientdecay0} and~\ref{lem:gradientdecay} respectively, we bound the rate of convergence of the gradient flow~\eqref{eq:gradientflow} stated in Theorem~\ref{thm:rateofconvergenceone}. The idea is that once the flow enters the approximate stationary state, it gets trapped there by the local g-convexity. Finally, under Assumption~\ref{assump:one}, we show a semi-monotonicity property of $\tracehess$ in Lemma~\ref{lem:closetoopt}, proved in \Cref{sec:proofclosetoopt}; namely, having small gradient at a point on the manifold implies that the neurons are close to an optimal point $\theta^*$, as defined in Equation~\eqref{eq:effectofdepth}, in the subspace spanned by data. This automatically translates our rate of decay of gradient in Lemma~\ref{lem:gradientdecay} to a convergence rate of $\theta(t)$ to $\theta^*$. 
    \\
    First, note that once the norm of gradient satisfies 
    \[
    \|\nabla \tracehess(\theta(t))\| \leq \epsilon
    \]
     for any $\epsilon \leq \sqrt \mu \beta$, it remains bounded by $\epsilon$ due to the non-increasing property of the norm of gradient guaranteed by Lemma~\ref{lem:gradientdecay0}. Therefore, if for some time $t$ we have $\|\nabla \tracehess(\theta(t))\| \leq \epsilon$ then $\|\nabla \tracehess(\theta(t'))\| \leq \epsilon$ for all $t' \geq t$. On the other hand, observe that the implicit regularizer decays with rate of squared norm of its gradient along the flow:
    \begin{align*}
        \frac{d\tracehess(\theta(t))}{dt} = -\|\nabla \tracehess(\theta(t))\|^2.
    \end{align*}
    Therefore, if for some time $t$ we have $\|\nabla \tracehess(\theta(t))\| \geq \epsilon$ we can lower bound the value of $\tracehess(\theta)$ at time $t$ as
    \begin{align*}
        0 \leq \tracehess(\theta(t)) &\leq \tracehess(\theta(0)) - \int_{t'=0}^t \|\nabla \tracehess(\theta(t'))\|^2 dt'\\
        &\leq  \tracehess(\theta(0)) - \int_{t'=0}^t \epsilon^2 dt'
        = \tracehess(\theta(0)) - t\epsilon^2,
    \end{align*}
    which implies
    \begin{align*}
        t \leq \tracehess(\theta(0))/\epsilon^2.
    \end{align*}
    To further strengthen this into an exponential convergence rate, we know once the norm of the gradient goes below $\sqrt \mu \beta$ it will decay exponentially fast due to Lemma~\ref{lem:gradientdecay}. From the first part, it takes at most $\tracehess(\theta(0))/(\mu \beta^2)$ time for the norm of gradient to go under threshold $\sqrt \mu \beta$, then after time $\log((\mu \beta^2/\epsilon'^2) \vee 1)/(\varrho_1\varrho_2 \mu)$ it goes below $\epsilon'$ using the exponential decay. Hence, the overall time $t_{\epsilon'}$ it takes so that norm of the gradient goes below threshold $\epsilon'$ is bounded by
    \begin{align}
        t_{\epsilon'}\leq \frac{\tracehess(\theta(0))}{\mu \beta^2} + \frac{\log((\mu \beta^2/\epsilon'^2) \vee 1)}{\varrho_1\varrho_2 \mu}.\label{eq:tepsilon}
    \end{align}

    Now we want to combine this result with Lemma~\ref{lem:closetoopt}. For the convenience of the reader, we restate Lemma~\ref{lem:closetoopt}.
    \begin{lemma*}[see Lemma~\ref{lem:closetoopt}]
        Suppose $\|\nabla \tracehess(\theta)\| \leq \delta$. Then, for all $i\in [n], j\in [m]$,
    \begin{align*}
        \Big|\theta_j^\top x_i - {\theta^*_j}^\top x_i\Big| \leq \delta/(\sqrt \mu \varrho_1 \varrho_2),
    \end{align*}
    for a $\theta^*$ as defined in Equation~\eqref{eq:effectofdepth}.
    \end{lemma*}

    Combining Equation~\eqref{eq:tepsilon} with Lemma~\ref{lem:closetoopt}, we find that to guarantee the property 
    \begin{align}
    |\theta^\top(t)_j x_i - {\theta^*_j}^\top x_i| \leq \epsilon, \forall i\in [n] , j \in [m],~\label{eq:dotproductclose}   
    \end{align}
     we need to pick $\epsilon' = \epsilon \sqrt \mu \varrho_1 \varrho_2$, which implies that we have condition~\eqref{eq:dotproductclose} guaranteed for all times 
    \[
    t\geq \frac{\tracehess(\theta(0))}{\mu \beta^2} + \frac{\log( \beta^2/(\varrho_1^2\varrho_2^2\epsilon^2) \vee 1)}{\varrho_1\varrho_2 \mu}.
    \]



\section{Remaining proofs for characterizing the stationary points and convergence}

\subsection{Proof of Lemma~\ref{lem:tracehessformula}}
    For any $\theta \in \Rmd$ we can write
    \begin{align*}
        \hess(\theta) = 2\sum_{i=1}^n Df_i(\theta) Df_i(\theta)^\top + 2\sum_{i=1}^n D^2 f_i(\theta) (f_i(\theta) - y_i).
    \end{align*}
    But from $\loss(\theta) = y_i$ we get for every $i\in [n]$, $f_i(\theta) = 0$, which implies
    \begin{align}
        \hess = 2\sum_{i=1}^n Df_i(\theta) Df_i(\theta)^\top.\label{eq:hessianformulaa}
    \end{align}
    Taking trace from both sides completes the proof.

\subsection{Proof of Lemma~\ref{lem:tracehesstwolayer}}\label{sec:prooflem2}
    Note that the gradient of the neural network function calculated on $x_i$ with respect to the parameter $\theta_j$ of the $j$th neuron is
    \begin{align*}
      D_{\theta_j} f_i(\theta) = \phi'(\theta_j^\top x_i)x_i.  
    \end{align*}
This implies
\begin{align}
    \|Df_i(\theta)\|^2 = \sum_{j=1}^m \phi'(\theta_j^\top x_i)^2.\label{eq:gradientinorm}
\end{align}
Summing Equations~\eqref{eq:gradientinorm} for all $j\in [m]$ completes the proof.

\subsection{Proof of Lemma~\ref{lem:hessianformula}}\label{sec:prooflem5}
Combining Facts~\ref{fact:covariant} and~\ref{fact:hessian} implies
\begin{align*}
    \nabla^2 F(\theta)[u,w] &= \Big\langle \nabla_u(\nabla F(\theta)), w\Big\rangle\\
    & =\Big\langle \thetaproj\Big(D(\thetaproj(DF(\theta)))[u]\Big), w\Big\rangle\\
    & \stackrel{\eqref{eq:mylabel}}{=}\Big\langle D(\thetaproj(DF(\theta)))[u], w\Big\rangle\numberthis\label{eq:mylabel}\\
    &=\Big\langle D\Big((I - \nproj)(DF(\theta))\Big)[u], w\Big\rangle,\numberthis\label{eq:hessianformula}
\end{align*}
where Equality~\eqref{eq:mylabel} follows because $w \in \tangentspace{\theta}$, the fact that for any vector $v$,
\[
v = \thetaproj(v) + \nproj(v),
\]
and that the part $\nproj\Big(D(\thetaproj(DF(\theta)))(u)\Big)$ has zero dot product with $w$.
Now note that from Lemma~\ref{lem:normalbasis}, the rows of the Jacobian matrix $Df(\theta)$ spans the normal space $\normalspace{\theta}$ for any $\theta \in \manifold$. Therefore, the projection matrix onto the normal space $\normalspace{\theta}$ at point $\theta \in \manifold$ regarding the operator $\nproj$ is given by
\begin{align}
    \nproj(v) = {Df(\theta)}^\top(Df(\theta){Df(\theta)}^\top)^{-1}Df(\theta)v.\label{eq:projectionformula}
\end{align}
Plugging Equation~\eqref{eq:projectionformula} into Equation~\eqref{eq:hessianformula}, we get:
\begin{align*}
    \nabla^2 F(\theta)[u,w] &= \Big\langle D\Big((I - {Df(\theta)}^\top(Df(\theta){Df(\theta)}^\top)^{-1}Df(\theta))DF(\theta)\Big)[u] , w\Big\rangle\\
    &= w^\top D\Big((I - {Df}^\top(Df{Df}^\top)^{-1}Df)DF\Big)[u]\\
    &= w^\top (I - {Df(\theta)}^\top(Df(\theta){Df(\theta)}^\top)^{-1}Df(\theta)) D^2F(\theta) u \\
    &+ w^\top D\Big(I - {Df(\theta)}^\top(Df(\theta){Df(\theta)}^\top)^{-1}Df(\theta)\Big)[u] DF(\theta),\numberthis\label{eq:plugback}
\end{align*}
where in the last line we used the chain rule. But note that $w \in \tangentspace{\theta}$, which implies $\thetaproj(w) = w$. This means 
\[
(I - {Df(\theta)}^\top(Df(\theta){Df(\theta)}^\top)^{-1}Df(\theta))w = w.
\]
Plugging this back into Equation~\eqref{eq:plugback} and noting the fact that $D(I)[u] = 0$,
\begin{align}
    & \nabla^2 F(\theta)[u,w] = w^\top D^2 F(\theta) u - w^\top D\Big( {Df(\theta)}^\top(Df(\theta){Df(\theta)}^\top)^{-1}Df(\theta)\Big)[u] DF(\theta).\label{eq:secondterm}
\end{align}
Now regarding the second term in Equation~\eqref{eq:secondterm}, note that the directional derivative in direction $u$ can either hit the $Df(\theta)$ terms or the middle part $(Df(\theta)Df(\theta)^\top)^{-1}$, i.e. we get
\begin{align*}
    w^\top D\Big( {Df(\theta)}^\top(Df(\theta){Df(\theta)}^\top)^{-1}Df(\theta)\Big)[u] &= 
    w^\top{D^2f(\theta)[u]}^\top(Df(\theta){Df(\theta)}^\top)^{-1}Df(\theta)\\
    &+ w^\top{Df(\theta)}^\top D\Big((Df(\theta){Df(\theta)}^\top)^{-1}Df(\theta)\Big)[u].\numberthis\label{eq:secondtermtwo}
\end{align*}
Now the key observation here is that because $w \in \tangentspace{\theta}$, we have
\[
Df(\theta)w = 0,
\]
which means the second term in Equation~\eqref{eq:secondtermtwo} is zero. Plugging this back into Equation~\eqref{eq:secondterm} implies
\begin{align*}
    \nabla^2 F(\theta)[u,w] = w^\top D^2 F(\theta) u - {D^2f(\theta)[u,w]}^\top (Df(\theta){Df(\theta)}^\top)^{-1}Df(\theta) DF(\theta),
\end{align*}
which completes the proof.

\subsection{Proof of Lemma~\ref{lem:hessstrongconvexity}}\label{sec:proofhessstrongconvexity}
    Let $\mathcal N$ be the subspace of $\tangentspace{\theta}$ which can be represented by 
    \begin{align}
    v = \stack{\sum_{i=1}^n \nu^j_i x_i}_{j=1}^m,\label{eq:subspaceform}    
    \end{align}
     for arbitrary coefficients $(\nu^j_i)_{i=1,\dots,n, j=1,\dots, m}$.
    First, we show that $\nabla \tracehess(\theta)\in \mathcal N$.
    Recall the definition of the gradient of $\tracehess$ on $\manifold$, i.e. 
    \[
    \nabla \tracehess(\theta) = \thetaproj(D\tracehess(\theta)) = D\tracehess(\theta) - \nproj(D\tracehess(\theta)).
    \]
    Note that $\nproj(D\tracehess(\theta)) \in \mathcal N$. This is because from Lemma~\ref{lem:normalbasis} we know that $\brackets{Df_i(\theta)}_{i=1}^n$ is a basis for $\normalspace{\theta}$ and each $Df_i(\theta)$ is clearly in the form~\eqref{eq:subspaceform}. Now we argue for any vector $v \in \mathcal N$, we have 
    \begin{align}
    \nabla^2 \tracehess(\theta)[v,v] \geq \mu\varrho_1\varrho_2\|v\|^2.\label{eq:key}    
    \end{align}
     Let $v \triangleq \stack{\sum_{i=1}^n \nu^j_i x_i}_{j=1}^m$. Note that for all $j \in [m]$, from Assumptions~\ref{assump:one} and~\ref{assump:two}:
     \begin{align*}
         \nabla^2 \tracehess(\theta)[v,v] &= \sum_{j=1}^m\sum_{i=1}^n \phi'(\theta_j^\top x_i) \phi'''(\theta_j^\top x_i)\Big((\sum_{i'=1}^n \nu^j_{i'} x_{i'})^\top x_i\Big)^2\\
         &\geq \varrho_1\varrho_2 \sum_{j=1}^m\sum_{i=1}^n\Big((\sum_{i'=1}^n \nu^j_{i'} x_{i'})^\top x_i\Big)^2\\
         &\geq \varrho_1\varrho_2 \mu \sum_{j=1}^m\|\sum_{i'=1}^n \nu^j_{i'} x_{i'}\|^2,
     \end{align*}
    where in the last inequality, we used the fact that from Assumption~\ref{assump:two}, for any vector $x$ in the subspace spanned by $\{x_i\}_{i=1}^n$ we have $x^\top XX^\top x \geq \mu \|x\|^2$. Combining this with Equation~\eqref{eq:hessianbound} in the proof of Lemma~\ref{lem:psdness} completes the proof.


Next, we prove Lemma~\ref{lem:closetoopt} which translates the approximate stationary property into topological closeness to a global optimum.

\subsection{Proof of Lemma~\ref{lem:closetoopt}}\label{sec:proofclosetoopt}
    Similar to the proof of Lemma~\ref{lem:psdness}, we have~\eqref{eq:alphaiprimeupperbound}:
    \begin{align*}
        |\phitwoprime - \alpha_i'| \varrho_1 \leq|(\phitwoprime - \alpha_i') \phiprime| \leq \delta/(2\sqrt{\mu}),
    \end{align*}
    which implies
    \begin{align}
        |\phitwoprime - \alpha_i'| \leq \delta/(2\sqrt \mu \varrho_1).\label{eq:tmp1}
    \end{align}
    Now since $\phi''' > 0$, we have that $\phi''$ is strictly monotone and invertible. Define
    \begin{align}
        \nu'_i = {\phi''}^{-1}(\alpha_i').\label{eq:tmp2}
    \end{align}
     The fact that $\phi''' \geq \varrho_2$ provides us with a Lipschitz constant of $1/\varrho_2$ for ${\phi''}^{-1}$. Using this with Equations~\eqref{eq:tmp1}  and~\eqref{eq:tmp2} implies
    \begin{align}
        \Big|\theta_j^\top x_i - \nu'_i\Big| \leq \delta/(2\sqrt \mu \varrho_1\varrho_2).\label{eq:nuprimeeq}
    \end{align}
    From Equation~\eqref{eq:nuprimeeq} we want to show $|\nu_i - \nu'_i| \leq \delta/(2\sqrt \mu \varrho_1 \varrho_2)$. Suppose this is not true. Then either $\nu_i > \nu'_i + \delta/(2\sqrt \mu \varrho_1\varrho_2)$ or $\nu_i < \nu'_i - \delta/(2\sqrt \mu \varrho_1\varrho_2)$. In the first case, using Equation~\eqref{eq:nuprimeeq} we get for all $j\in [m]$:
    \begin{align*}
        \nu_i > \theta_j^\top x_i,
    \end{align*}
    which from the strict monotonicity of $\phi$ implies
    \begin{align*}
        y_i/m = \phi(\nu_i) > \phi(\theta_j^\top x_i),   
    \end{align*}
    which means $f_i(\theta) < y_i$. But this clearly contradicts with the fact that $\theta$ is on the manifold of zero loss, i.e. $f_i(\theta) = y_i$. In the other case $\nu_i < \nu'_i - \delta/(2\sqrt \mu \varrho_1\varrho_2)$ we can get a similar contradiction. Hence, overall we proved
    \begin{align*}
        |\nu_i - \nu'_i| \leq \delta/(2\sqrt \mu \varrho_1 \varrho_2).
    \end{align*}
    Combining this with Equation~\eqref{eq:nuprimeeq} completes the proof.

\section{Proof of convergence of the flow to $\manifold$}
In this section, we analyze the behavior of label noise SGD starting from $\theta_0$ with positive loss, when step size goes to zero. To this end, we study a different gradient flow than the one in Equation~\eqref{eq:gradientflow}.
\subsection{Gradient flow regarding label noise SGD in the limit}
In Lemma~\ref{lem:firtflowconverge}, we prove that under Assumptions~\ref{assump:one},~\ref{assump:two}, the gradient flow with respect to the gradient $-D\loss(\theta)$ converges to the manifold $\manifold$ exponentially fast. This is in particular important to in the proof of Theorem~\ref{thm:main_intro}, in particular in proving that SGD with small enough step size will reach zero loss, because SGD in the limit of step size going to zero will converge to this gradient flow outside of $\manifold$. The key to show this result is a PL inequality that we prove in this setting for $\loss$, in Lemma~\ref{lem:plinequality}. Here we obtain explicit constants which is not necessary for proving Theorem~\ref{thm:main_intro}.
\begin{lemma}[Convergence of the gradient flow to the manifold]\label{lem:firtflowconverge}
    In the limit of step size going to zero, label noise SGD initialized at a point $\thetazero$ converges to the following gradient flow 
    \begin{align}
     &\bar \theta(0) = \thetazero,\\
     &\bar \theta'(t) = -D\loss(\bar \theta(t)).\label{eq:firstgradientflow}
    \end{align}
   Under Assumption~\ref{assump:one} and~\ref{assump:two}, after time 
   \[
   t\geq \log(1/\epsilon)
   \]
   we have,
   \begin{align*}
       &\loss(\bar \theta(t)) \leq e^{-C t}\loss(\barthetazero),
   \end{align*}
   for constant $C = 4m\mu \varrho_1^2$. Furthermore, $\theta(t)$ converges to $\tilde \theta \in \manifold$ such that for all $t > 0$
   \begin{align}
       \|\bar\theta(t) - \tilde \theta\| \leq \frac{2}{\sqrt C} e^{-Ct/2}\sqrt{\loss(\barthetazero)},\label{eq:disttoinit}
   \end{align}
   and 
   \begin{align*}
       \tracehess(\tilde \theta) &\leq \sum_{i=1}^n \sum_{j=1}^m {\phi'}^2({{\thetazero}_j}^\top x_i) + \sum_{i=1}^n \max_{j=1}^m\{{\phi'}^2({{\thetazero}_j}^\top x_i) \pm \frac{2}{\sqrt C}\sqrt{\loss(\thetazero)}\}\\
       &\leq 2\tracehess(\theta_0)+ \frac{2}{\sqrt C}\sqrt{\loss(\theta_0)}.
   \end{align*}
\end{lemma}
\subsection{Proof of Lemma~\ref{lem:firtflowconverge}}
    First, taking derivative from $\loss(\bar \theta(t))$ and using the PL inequality from Lemma~\ref{lem:plinequality}:
    \begin{align*}
        \frac{d}{dt}\loss(\bar \theta(t)) = -\|D\loss(\bar \theta)\|^2 \leq -4m\mu\varrho_1^2 \loss(\bar \theta),
    \end{align*}
    which implies
    \begin{align}
        \loss(\bar \theta(t)) \leq e^{-4m\mu\varrho_1^2 t}\loss(\thetazero).\label{eq:exponentialdec}
    \end{align}
    
    Moreover, similar to page 38 in~\cite{li2021happens}, given the PL constant $C = 4m\mu \varrho_1^2$ we have
    \begin{align*}
        \Big\|\frac{d\bar \theta(t)}{dt}\Big\| = \Big\|\nabla \loss(\bar \theta(t))\Big\|
        \leq \frac{\Big\|\nabla \loss(\bar \theta(t))\Big\|^2}{\sqrt{C \loss(\bar \theta(t))}} = \frac{-\frac{d\loss(\bar \theta(t))}{dt}}{\sqrt{C \loss(\bar \theta(t))}} = -\frac{2}{\sqrt C}\frac{d\sqrt{\loss(\bar \theta)}}{dt}.\numberthis\label{eq:pathlength}
    \end{align*}
    This implies
    \begin{align*}
        \|\bar \theta(t) - \bar \theta(0)\| &\leq \int_{0}^t \Big\|\frac{d\bar \theta(t)}{dt}\Big\|dt \leq -\frac{2}{\sqrt C} \int_{0}^t \frac{d\sqrt{\loss(\bar \theta(t))}}{dt} \\
        &= \frac{2}{\sqrt C} (\sqrt{\loss(\bar \theta(0))} - \sqrt{\loss(\bar \theta(t))}) \leq \frac{2}{\sqrt C}\sqrt{\loss(\bar \theta(0))}.\numberthis\label{eq:distancebound}
    \end{align*}
    Additionally, from Equation~\eqref{eq:pathlength} we get for $t_1 \geq t_2$:
    \begin{align}
        \|\bar \theta(t_1) - \bar \theta(t_2)\| \leq \frac{2}{\sqrt C} (\sqrt{\loss(\bar \theta(t_1))}),\label{eq:t1t2}
    \end{align}
    which combined with Equation~\eqref{eq:exponentialdec} implies that $\theta(t)$ converges to some $\tilde \theta$. On the other hand, since $\loss(\theta(t))$  converges to $\loss(\tilde \theta)$ by continuity of $\loss$ and because $\loss(\bar \theta(t))$ is going to zero, we conclude that the limit point $\tilde \theta$ should have zero loss, i.e. $\tilde \theta \in \manifold$. Now applying Equation~\eqref{eq:t1t2} for $t_2 = t$ and sending $t_1$ to infinity implies for all $t > 0$:
    \begin{align*}
        \|\bar \theta(t) - \tilde \theta\| \leq \frac{2}{\sqrt C} \sqrt{\loss(\bar \theta(t))} \leq \frac{2}{\sqrt C} e^{-Ct/2}\sqrt{\loss(\thetazero)}.
    \end{align*}
     On the other hand, recall that from Equation, trace of Hessian on the manifold is equal to
    \begin{align}
        \tracehess(\theta) = \sum_{i=1}^n \Big\|\nabla f_i(\bar \theta)\Big\|^2 = \sum_{i=1}^n \sum_{j=1}^m \phi'({\bar \theta_j}^\top x_i)^2.\label{eq:tracehesstwo}
    \end{align}
    But under Assumption~\ref{assump:one} we have that $\phi'^2$ is convex, because
    \begin{align*}
        \frac{d^2}{dz^2}\phi'(z)^2 = 2\phi''(z)^2 + 2\phi'(z)\phi'''(z) \geq 0.
    \end{align*}
    But from Equation~\eqref{eq:distancebound} we have
    \begin{align*}
        \sum_{j=1}^m\big|\bar{\theta}^\top_j(t) x_i - \bar{\theta}^\top_j(0) x_i\big| \leq \|\bar \theta(t) - \bar \theta(0)\| \leq \frac{2}{\sqrt C}\sqrt{\loss(\bar \theta(0))}.
    \end{align*}
    Combining the convexity and positivity of $\phi'^2$ with Equation~\eqref{eq:distancebound} and the formula for $\Big\|\nabla f_i(\theta)\Big\|^2$:
    \begin{align*}
        \Big|\Big\|\nabla f_i(\bar \theta(t))\Big\|^2 - \Big\|\nabla f_i(\bar \theta(0))\Big\|^2\Big| &\leq \max_{j=1}^m\{\phi'^2(\bar \theta^\top_j(0) x_i \pm \frac{2}{\sqrt C}\sqrt{\loss(\bar \theta(0))}) - \phi'^2(\bar{\theta}^\top_j(0) x_i)\}\\
        &\leq \max_{j=1}^m\{\phi'^2(\theta^\top_j(0) x_i \pm \frac{2}{\sqrt C}\sqrt{\loss(\bar \theta(0))})\},\numberthis\label{eq:diffnorm}
    \end{align*}
    where $\pm$ above in the argument of $\phi'$ means we take maximum with respect to both $+$ and $-$. Summing Equation~\eqref{eq:diffnorm} for all $i$, we have for constant
    \[
    C_2 \coloneqq \sum_{i=1}^n \max_{j=1}^m\{\phi'^2(\bar \theta^\top_j(0) x_i \pm \frac{2}{\sqrt C}\sqrt{\loss(\bar \theta(0))})\},
    \]
    we have
    \begin{align*}
        \sum_{i=1}^n \Big\|\nabla f_i(\bar \theta(t))\Big\|^2 \leq \sum_{i=1}^n \Big\|\nabla f_i(\thetazero)\Big\|^2 + C_2.
    \end{align*}
    Sending $t$ to infinity implies
    \begin{align*}
        \sum_{i=1}^n \Big\|\nabla f_i(\tilde \theta)\Big\|^2 \leq \sum_{i=1}^n \Big\|\nabla f_i(\thetazero)\Big\|^2 + C_2.
    \end{align*}
    But since $\tilde \theta \in \manifold$, from~\eqref{eq:tracehesstwo} we have
    \begin{align*}
        \tracehess(\tilde \theta) \leq \sum_{i=1}^n\sum_{j=1}^m \phi'({{\thetazero}_j}^\top x_i)^2 + C_2,
    \end{align*}
    which completes the proof.

\subsection{A PL inequality for $\loss$}
Next, in Lemma~\ref{lem:plinequality} we show a PL inequality for $\loss$ under Assumptions~\ref{assump:one} and~\ref{assump:two}, which we then use to show Lemma~\ref{lem:firtflowconverge}.
\begin{lemma}[PL inequality outside of manifold]\label{lem:plinequality}
    Under Assumptions~\ref{assump:one} and~\ref{assump:two}, the loss $\loss(\theta)$ satisfies a PL inequality of the form
    \begin{align*}
        \|D\loss(\bar \theta)\|^2 \geq 4m\mu \varrho_1^2 \loss(\bar \theta).
    \end{align*}
\end{lemma}
\begin{proof}
    Note that for all $j\in [m]$, $D_{\theta_j}\loss(\bar \theta)$ is given by
    \[
    D_{\bar{\theta}_j}\loss(\bar \theta) = 2\sum_{i=1}^n (r_{\bar \theta, NN}(x_i) - y_i) \phi'(\bar{\theta}_j^\top x_i)x_i.
    \]
    Hence
    \begin{align*}
        \|D_{\bar{\theta}_j}\loss(\bar \theta)\|^2 &= 4\Big\|\sum_{i=1}^n (r_{\bar \theta, NN}(x_i) - y_i) \phi'(\bar{\theta}_j^\top x_i)x_i\Big\|^2 \\
        &\geq 4\mu \sum_{i=1}^n \phi'(\bar{\theta}_j^\top x_i)^2 (r_{\bar{\theta}, NN}(x_i) - y_i)^2\\
        & \geq 4\mu\varrho_1^2 \loss(\bar \theta).\numberthis\label{eq:subpl}
    \end{align*}
    summing Equation~\eqref{eq:subpl} for all neurons completes the proof.
\end{proof}
\section{Proof of the final guarantee for Label noise SGD}
In this section we prove Theorem~\ref{thm:main_intro}.
\subsection{Proof of \Cref{thm:main_intro}}\label{sec:proofintrothm}
    Our main theorem (\Cref{thm:main_intro}) is a direct combination of Theorem 4.6 of \citet{li2021happens}, and \Cref{thm:stationary} and \Cref{thm:rateofconvergenceone}.
    First note that from Lemma~\ref{lem:firtflowconverge}, the gradient flow in the limit of step size going to zero always converges to the manifold of zero, independent of the initialization. 
    Therefore, the neighborhood $U$ in Theorem 4.6 of~\cite{li2021happens} is the whole $\Rr$, which implies that for step size $\eta < \eta_0$ less than a certain threshold $\eta_0$, $\theta_{\lceil t/\eta^2 \rceil}$ is close enough to $\theta(t)$ in distribution, where $\theta(t)$ is
    the Riemannian gradient flow defined in Theorem~\ref{thm:rateofconvergenceone} which is initialized as $\theta(0) = \tilde \theta$. Recall that $\tilde \theta$ is the limit point of the first gradient flow (before reaching the manifold $\manifold$) defined  in Lemma~\ref{lem:firtflowconverge}.
    
    Moreover, again from Lemma~\ref{lem:firtflowconverge}, the gradient flow $\bar \theta(t)$ defined in that Lemma remains within distance of $O(\frac{2}{\sqrt C} \sqrt{\loss(\barthetazero)})$ of initialization. Therefore, trace of Hessian of the loss at $\tilde \theta$ is bounded and only depends on the initialization $\theta_0$. This enables us to have a bounded initial value for the second ODE in~\eqref{eq:gradientflow}, depending only on the primary initialization $\bar \theta(0)$ (which can be outside the manifold). 
    In particular, note that from Theorems~\ref{thm:stationary} and~\ref{thm:rateofconvergenceone}, this flow converges to $\epsilon'$-proximity of a global minimizer $\theta^*$ of trace of Hessian on $\manifold$ in the sense of~\eqref{eq:dotproductproximity} after time at most $\tilde O(\log(1/\epsilon))$.
    
    Now using the unit norm assumption $\|x_i\| = 1$ and picking the step size threshold $\eta_0$ small enough, we see that after at most $K = \Theta((1 + \log(1/\epsilon'))/\eta^2)$ iterations of label noise SGD for $\eta \leq \eta_0$, for all $j\in [m]$ and $i \in [n]$ we have with high probability
    \begin{align}
        \big|{\theta_{T}}_j^\top x_i - {\theta^*}_j^\top x_i\big| \leq 2\epsilon'.\label{eq:guarantee}
    \end{align}
    Therefore, picking $\epsilon' \leq \epsilon/2$, we get
    for all $i\in [n]$ and $j\in [m]$:
    \begin{align*}
        \big|{\theta_T}_j^\top x_i - \phi^{-1}(y_i/m)\big| \leq \epsilon.
    \end{align*}
    Moreover, note that from Equation~\eqref{eq:disttoinit} and Lemma~\ref{lem:boundedregion}, $\theta$ remains in a ball of bounded radius depending only on $\theta_0$, hence $\loss$ and $\tracehess$ both have a bounded Lipschitz constant only depending on the initialization as well. Therefore, we pick $\epsilon'$ small enough so  that we get $\loss(\theta_K) \leq \epsilon$ and
    $\Big|\tracehess(\theta_K) - \tracehess(\theta^*)\Big| \leq \epsilon$
    for the final iteration $K$ of label noise SGD.
    This completes the proof.
\section{Derivative formulas}
\subsection{Derivative tensors}
In Lemma~\ref{lem:derivativeequations} we calculate the Jacobian of $f$ and the Hessian of $f_i$'s.
\begin{lemma}\label{lem:derivativeequations}
The Jacobian of $f$ is given by:
\begin{align*}
    Df = 
    \begin{bmatrix}
    \phi'(\theta_1^\top x_1)x_1^\top & \dots & \dots \\
    \dots & \phi'(\theta_j^\top x_i)x_i^\top, \ \phi'(\theta_{j+1}^\top x_{i})x_{i}^\top  & \dots \\
    \dots & \dots & \dots. \\
    \end{bmatrix}
\end{align*}
Moreover, the Hessian of $f_i$'s is given by
\begin{align*}
    D^2f_i(.,.) = 
    \begin{bmatrix}
    \phi''(x_i^\top \theta_1) x_ix_i^\top & 0 & \dots\\
    0\dots & \dots\\
    0\dots & \phi''(x_i^\top \theta_j) x_ix_i^\top & 0\dots\\
    & \dots &.
    \end{bmatrix}  
\end{align*}
As a result
\begin{align*}
    D^2f_i(u,u) = \sum_{j=1}^N \phi''(x_i^\top \theta_j) (x_i^\top u_j)^2.
\end{align*}
\end{lemma}

In this Lemma~\ref{lem:derivativeequation2}, we calculate the Hessian of the trace of Hessian regularizer in our two-layer setting.
\begin{lemma}\label{lem:derivativeequation2}
For the implicit regularizer $F \triangleq \tracehess$ regarding our two-layer network with mean squared loss as in~\eqref{eq:meansquaredloss}, we have
\begin{align*}
    F(\theta) = \sum_{i=1}^n \sum_{j=1}^m \phi'^2(\theta_j^\top x_i)\|x_i\|^2 = \sum_i \sum_j \phi'^2(\theta_j^\top x_i),
\end{align*}
and
\begin{align}\label{eq:euclideanhessian} 
    &D^2F(.,.) \\
    &= 
    \sum_i 
    \begin{bmatrix}
    (2{\phi''}^2(x_i^\top \theta_1) + \phi'''(x_i^\top \theta_1)\phi'(x_i^\top \theta_1)) x_ix_i^\top & 0 & \dots\\
    0\dots & \dots
    0\dots & (2{\phi''}^2(x_i^\top \theta_j) + \phi'''(x_i^\top \theta_j)\phi'(x_i^\top \theta_j)) x_ix_i^\top & 0\dots\\
    & \dots &.
    \end{bmatrix}
\end{align}
Hence
\begin{align}
    D^2F(\theta)[u,u] = \sum_{i=1}^n\sum_{j=1}^m(2\phi''^2(x_i^\top \theta_j) + 2\phi'''(x_i^\top \theta_j)\phi'(x_i^\top \theta_j))\langle x_i, u_j\rangle^2.\label{eq:hessianmainterm}
\end{align}
\end{lemma}

\section{Other proofs}
In Lemma~\ref{lem:boundedregion} we show that the gradient flow in Equation~\eqref{eq:gradientflow} remains in a bounded region.
\begin{lemma}[Bounded region]\label{lem:boundedregion}
    Under the positivity assumption $\phi''' > 0$, the gradient flow $\frac{d}{dt}\theta(t) = -\nabla \tracehess(\theta(t))$ remains in a bounded region for all times $t$.
\end{lemma}
\begin{proof}
    Note that the value of $\tracehess$ is decreasing along the flow, so for all $t\geq 0$:
    \begin{align*}
        \tracehess(\theta(t)) \leq \tracehess(\thetazero).
    \end{align*}
    But noting the formula of $\tracehess$ in Lemma~\ref{lem:tracehesstwolayer}, we get that for all $j\in [m]$ and $i\in [n]$:
    \begin{align}
    \phi'(\theta_j^\top x_i)^2 \leq \tracehess(\thetazero).\label{eq:activationbound}
    \end{align}
    Now let $z^*$ be the minimizer of $\phi'$. Now from the assumption of the Lemma we have $\phi'''(z^*) > 0$, which implies from continuity of $\phi'''$ that for $\epsilon', \delta > 0$ we have for all $z\in (z^*-\epsilon', z^*+\epsilon')$, 
    \[
    \phi'''(z) \geq \delta'.
    \]
    This means for all $z \leq z^*-\epsilon'$ we have $\phi''(z) \leq -\epsilon'^2\delta'/2$ and for $z \geq z^* + \epsilon'$ we have $\phi'(z) \geq \epsilon'^2\delta'/2$. Even more, for any $|z - z^*| \geq \epsilon'$ we have
    \begin{align*}
        \phi'(z) \geq (|z-z^*| - \epsilon')\epsilon'\delta' + \epsilon'^2\delta'/2 = (|z-z^*| - \epsilon'/2)\epsilon'\delta'.
    \end{align*}
    Therefore, Equation~\eqref{eq:activationbound} implies for $z = \theta_j^\top x_i$:
    \begin{align*}
        (|z-z^*| - \epsilon'/2)\epsilon'\delta' \leq \tracehess(\thetazero),
    \end{align*}
    or
    \begin{align*}
        |z - z^*| \leq \tracehess(\thetazero)/(\epsilon'\delta') + \epsilon'/2.
    \end{align*}
    This implies the boundedness of the arguments of all the activations for all data points, which completes the proof.
\end{proof}

\begin{lemma}[$\mathcal M$ is well-defined]\label{lem:manifoldwelldefined}
    The intersection of the sub-level sets $f_i(\theta) = y_i$, defined in Section~\ref{sec:proofsketches} as $\mathcal M$, is well-defined as a differentiable manifold.
\end{lemma}
\begin{proof}
    Due to a classical application of the implicit function theorem after a linear change of coordinates using a basis for the kernel of the Jacobian, it turns out that there exists a local chart for $\mathcal M$ around each point $\theta \in \mathcal M$ for which the Jacobian matrix is non-degenerate. But due to Lemma~\ref{lem:nonsingular} the Jacobian is always non-degenerate, so $\mathcal M$ is indeed well-defined as a manifold.
\end{proof}

\begin{lemma}[Non-singularity of the Jacobian]\label{lem:nonsingular}
    Under assumption~\ref{assump:one}, the Jacobian $Df$ is non-singular, i.e. its rank is equal to the number of its rows.
\end{lemma}
\begin{proof}
    For an arbitrary data point $\theta_j$, we show the non-singularity of the submatrix of $Df$ whose columns corresponds to $\theta_j$, namely $D_{\theta_j}f$. Any linear combination of the rows of this matrix is of the form
    \begin{align}
        \sum_{i=1}^n \alpha_i \phi'(\theta_j^\top x_i) x_i.\label{eq:linearcomb}
    \end{align}
    But from the coherence assumption in~\ref{assump:one}, we see that $\{x_i\}_{i=1}^n$ are linearly independent, so the linear combination in Equation~\eqref{eq:linearcomb} is also non-zero, which implies the non-singularity of $D_{\theta_i}f$ and completes the proof.
\end{proof}

\begin{lemma}[Cube activation]\label{lem:cube}
    If the distribution of the labels does not have a point mass on zero (i.e. if $\mathbb P(y = 0) = 0$), then for the cube activation the manifold $\mathcal M$ is well-defined, and Theorem 2 also holds.
\end{lemma}
\begin{proof}
    To show that $\mathcal M$ is well-defined, similar to Lemma~\ref{lem:manifoldwelldefined} we show the jacobian $Df(\theta)$ is non-degenerate for all $\theta$. To this end, it is enough to show that for each $x_i$, there is a $j$ such that $D_{\theta_j}f_i$ is non zero. This is because $D_{\theta_j}f_i$ is always a scaling of $x_i$, so if it is non-zero, then it is a non-zero scaling of $x_i$, and since $x_i$'s are linearly independent from Assumption~\ref{assump:one} the non-degeneracy of $Df$ follows. To show the aformentioned claim, note that for each $\theta \in \mathcal M$, from the assumption on the distribution of the label, we have with probability one:
    \begin{align*}
        \sum_{j=1}^m \phi({\theta_j}^\top x_i) = y_i \neq 0.
    \end{align*}
    Therefore, there exists $\tilde j$ such that
    \begin{align*}
        \phi({\theta_j}^\top x_i) \neq 0,
    \end{align*}
    which implies 
    \begin{align*}
        {\theta_j}^\top x_i \neq 0,
    \end{align*}
    which means
    \begin{align*}
        \phi'({\theta_j}^\top x_i) \neq 0.
    \end{align*}
    Therefore,
    \begin{align*}
        D_{\theta_j} f_i(\theta) = \phi'({\theta_j}^\top x_i)x_i \neq 0,
    \end{align*}
    and the proof of the non-degeneracy of $Df$ is complete.

    Next, we show that proof of Theorem~\ref{thm:stationary} is still valid even though Assumption~\ref{assump:one} does not hold for $\phi(x) = x^3$. This is because the only argument that we use in the proof of Theorem~\ref{thm:stationary} which depends on Assumption~\ref{assump:one} is the invertibility of $\phi$ and $\phi''$ which holds for the cube activation. Hence, the proof of complete.
    
\end{proof} 
\section{G-convexity}
In the following Lemma~\ref{lem:localgconvexity}, we show that PSD property of the Hessian of an arbitrary smooth function $F$ on $\manifold$ at point $\theta$ implies that its locally g-convex at $\theta$.
\begin{lemma}[Local g-convexity]\label{lem:localgconvexity}
For a function $F$ on manifold $\manifold$, given that its Hessian is PSD in an open neighborhood $\mathcal C \subseteq \manifold$, then $F$ is $g$-convex in $\mathcal C$ on $\manifold$.
\end{lemma}
\begin{proof}
    Consider a geodesic $\gamma(t) \subseteq \mathcal M$. Then taking derivative of $F$ on $\gamma$ with respect to $t$:
    \begin{align*}
        &\frac{d}{dt}F(\gamma(t)) = \langle \nabla F(\gamma(t)), \gamma'(t)\rangle,
    \end{align*}
    and for the second derivative
    \begin{align*}
        \frac{d^2}{dt^2}F(\gamma(t))& = \langle \nabla F(\gamma(t)), \nabla_{\gamma'(t)}\gamma'(t)\rangle + \langle \nabla^2 F(\gamma(t))\gamma'(t), \gamma'(t)\rangle\\
        &= \langle \nabla F(\gamma(t)), \nabla_{\gamma'(t)}\gamma'(t)\rangle,
    \end{align*}
    where we used the fact that $\nabla_{\gamma'(t)}\gamma'(t) = 0$ for a geodesic, and the fact that the Hessian of $F$ is PSD. Therefore, $F$ is convex in the normal sense on any geodesic in $\mathcal C$, which means it is g-convex in $\mathcal C$.
\end{proof}
\section{Background on Riemannian Geometry}\label{sec:rgeometry}
In this section, we recall some of basic concepts in differential geometry, related to our analysis in this work. The material here is directly borrowed from~\cite{gatmiry2022convergence,gatmiry2023sampling}. For a more detailed treatment on Differential Geometry, we refer the reader to~\cite{do2016differential}.

\paragraph{Abstract manifold.}
 A manifold $\mathcal M$ is a topological space such that for an integer $n\in\mathbb{N}$ and given each point $p \in \mathcal M$, there exists an open set $U$ around $p$ such that $U$ is a homeomorphism to an open set of  $\mathbb R^n$. We call a function $f$ defined on $\mathcal M$ smooth if for every $p\in \mathcal M$ and open set $U$ around $p$ which is homeomorphism to an open set $V \subseteq \mathbb R^n$ with mapping $\phi$, then $f \circ \phi$ is a smooth function of $V$. This is how we define vector fields on $\mathcal M$ later on.

Note that the manifold we consider in this work is a sub-manifold embedded in the Euclidean space, given by the intersection of sublevel sets of $f_i(\theta)$'s. Given that the Jacobian of $f = (f_i)_{i=1}^n$ is non-singular(\Cref{lem:nonsingular}), the manifold $\manifold = \{x \mid \ \forall i\in [n]: \ f_i(x) = y_i\}$ is well-defined.

\paragraph{Tangent space.} For any point $p \in \mathcal M$, one can define the notion of tangent space for $p$, $T_p(\mathcal M)$, as the set of equivalence class of differentiable curves $\gamma$ starting from $p$ ($\gamma(0)= p$), where we define two such curves $\gamma_0$ and $\gamma_1$ to be equivalent if for any function $f$ on the manifold: 
    \begin{align*}
        \frac{d}{dt}f(\gamma_0(t))\big|_{t=0} = \frac{d}{dt}f(\gamma_1(t))\big|_{t=0}.
    \end{align*}
When the manifold is embedded in $\mathbb R^d$, the tangent space at $p$ can be identified by the subspace of $\mathbb R^d$ obtained from the tangent vectors to curves on $\mathcal M$ passing through $p$.

\paragraph{Vector field.} 
A vector field $V$ is a smooth choice of a vector $V(p) \in T_p(\mathcal M)$ in the tangent space for all $p \in \mathcal M$.  

\paragraph{Metric and inner product.} 
A metric $\langle ., .\rangle$ is a tensor  
on the manifold $\mathcal M$ that is simply a smooth choice of a symmetric bilinear map on the tangent space of each point $p \in \mathcal M$. 
Namely, for all $w,v,z\in T_p(\mathcal M)$:
\begin{align*}
    &\langle v+w, z\rangle = \langle v,z\rangle + \langle w, z\rangle,\\
    &\langle \alpha v, \beta w\rangle =
    \alpha \beta\langle v, w\rangle.
\end{align*}

\paragraph{Covariant derivative.}
Given two vector fields $V$ and $W$, the covariant derivative, also called the Levi-Civita connection $\nabla_{V}W$ is a bilinear operator with the following properties:
\begin{align*}
    \nabla_{\alpha_1 V_1 + \alpha_2V_2}W = \alpha_1\nabla_{V_1}W + \alpha_2\nabla_{V_2}W,\\
    \nabla_{V}(W_1 + W_2) = \nabla_V(W_1) + \nabla_V(W_2),\\
    \nabla_V(\alpha W_1) = \alpha \nabla_V(W_1) + V(\alpha)W_1
    \end{align*}
    where $V(\alpha)$ is the action of vector field $V$ on scalar function $\alpha$.
Importantly, the property that differentiates the covariant derivative from other kinds of derivaties over the manifold is that the covariant derivative of the metric is zero, i.e., 
$\nabla_V g = 0$ for any vector field $V$. In other words, we have the following intuitive rule:
\begin{align*}
    \nabla_V \langle W_1, W_2 \rangle 
    = \langle \nabla_V W_1, W_2\rangle + \langle  W_1, \nabla_V W_2\rangle.
\end{align*}
Moreover, the covariant derivative has the property of being torsion free, meaning that for vector fields $W_1, W_2$:
\begin{align*}
    \nabla_{W_1}W_2 - \nabla_{W_2}W_1 = [W_1, W_2],
\end{align*}
where $[W_1, W_2]$ is the Lie bracket of $W_1, W_2$ defined as the unique vector field that satisfies
\begin{align*}
    [W_1, W_2]f = W_1(W_2(f)) - W_2(W_1(f))
\end{align*}
for every smooth function $f$.

In a local chart with variable $x$, if one represent $V = \sum V^i \partial x_i$, where $\partial x_i$ are the basis vector fields, and $W = \sum W^i \partial x_i$, the covariant derivative is given by
\begin{align*}
    \nabla_V W & = \sum_i V^i\nabla_i W =
    \sum_i V^i \sum_j \nabla_i (W^j \partial x_j) \\
    & = \sum_i V^i \sum_j \partial_i(W^j) \partial x_j + \sum_i V^i \sum_j W^j \nabla_i \partial x_j\\
    & =  \sum_j V(W^j) \partial x_j + \sum_i \sum_j V^i W^j \sum_k \Gamma_{ij}^k \partial x_k
    = \\
    & = \sum_k \big(V(W^k) + \sum_i \sum_j V^i W^j \Gamma_{ij}^k \big)\partial x_k.\label{eq:covariantexpansion}
\end{align*}
The Christoffel symbols $\Gamma_{ij}^k$ are the representations of the Levi-Cevita derivatives of the basis $\{\partial x_i\}$:
\begin{align*}
    \nabla_{\partial x_j} \partial x_i = \sum_{k} \Gamma_{ij}^k \partial x_k
\end{align*}
and are given by the following formula:
\begin{align*}
    \Gamma_{ij}^k = \frac{1}{2}\sum_{m} g^{km}(\partial_j g_{mi} + \partial_i g_{mj} - \partial_m g_{ij}).
\end{align*}
Above, $g^{ij}$ refers to the $(i,j)$ entry of the inverse of the metric. 

\paragraph{Gradient.}
For a function $F$ over the manifold, it naturally acts linearly over the space of vector fields: given a vector field $V$, the mapping $V(F)$ is linear in $V$. Therefore, by Riesz representation theorem 
there is a vector field $W$ where
\begin{align*}
    V(F) = \langle W,V \rangle 
\end{align*}
for any vector field $V$. This vector field $W$ is defined as the gradient of $F$, which we denote here by $\nabla F$. To explicitly derive the form of gradient in a local chart, note that
\begin{align*}
    V(F) = (\sum_{i} \partial x_i V_i)(F) = \sum_i V_i \partial_i F = \sum_{j,m,i} V_j g_{jm} g^{mi}\partial_i F = \langle V, \big(\sum_{i}g^{mi} \partial_i F\big)_{m}\rangle.
\end{align*}
The above calculation implies
\begin{align*}
    \nabla F = \sum_m (\sum g^{mi} \partial x_i F)\partial x_m.
\end{align*}
So $ g^{-1}DF$ is the representation of $\nabla F$ in the Euclidean chart with basis $\{\partial x_i\}$.

    Given a smooth function $F$ on $\Rr$ and sub-manifold $\manifold \subseteq \Rr$, the gradient $\nabla F$ on $\manifold$ at any point $\theta \in \manifold$ is given by the Euclidean projection of the normal Euclidean gradient of $F$ at $\theta$ onto the tangent space $\tangentspace{\theta}$.

\end{document}